%% file: main.tex
\documentclass[10pt,twocolumn,letterpaper]{article}

\usepackage{cvpr}              
\usepackage{multirow}
\usepackage{arydshln}

\definecolor{cvprblue}{rgb}{0.21,0.49,0.74}
\usepackage[pagebackref,breaklinks,colorlinks,allcolors=cvprblue]{hyperref}

\def\paperID{14380} 
\def\confName{CVPR}
\def\confYear{2025}

\title{Fish-Vista: A Multi-Purpose Dataset for Understanding \& Identification of Traits from Images}

\author{Kazi Sajeed Mehrab\textsuperscript{1}, M. Maruf\textsuperscript{1}, Arka Daw\textsuperscript{3}, Abhilash Neog\textsuperscript{1}, Harish Babu Manogaran\textsuperscript{1},
\\
Mridul Khurana\textsuperscript{1}, Zhenyang Feng\textsuperscript{2}, Bahadir Altintas\textsuperscript{6}, Yasin Bakis\textsuperscript{6}, Elizabeth G Campolongo\textsuperscript{2},
\\
Matthew J Thompson\textsuperscript{2}, Xiaojun Wang\textsuperscript{6}, Hilmar Lapp\textsuperscript{5}, Tanya Berger-Wolf\textsuperscript{2}, Paula Mabee\textsuperscript{7}, Henry Bart\textsuperscript{6},
\\
Wei-Lun Chao\textsuperscript{2}, Wasila M Dahdul\textsuperscript{4}, Anuj Karpatne\textsuperscript{1}
\\
\\
\textsuperscript{1}Virginia Tech, \textsuperscript{2}The Ohio State University, \textsuperscript{3}Oak Ridge National Laboratory, 
\\
\textsuperscript{4}University of California, Irvine, \textsuperscript{5}Duke University, \textsuperscript{6}Tulane University, \textsuperscript{7}Battelle
}

\begin{document}

\maketitle

\input{sec/0_abstract}

\input{sec/1_introduction}
\input{sec/2_related_works}

\input{sec/3_methods}

\input{sec/4_results}

\input{sec/5_conclusion}



{
    \small
    \bibliographystyle{ieeenat_fullname}
    \bibliography{appendix_references}
}

\input{sec/X_suppl}

\end{document}

%% file: sec/0_abstract.tex
\begin{abstract}

We introduce Fish-Visual Trait Analysis (Fish-Vista), the first organismal image dataset designed for the analysis of visual traits of aquatic species directly from images using problem formulations in computer vision. Fish-Vista contains 69,126 annotated images spanning 4,154 fish species, curated and organized to serve three downstream tasks of species classification, trait identification, and trait segmentation. Our work makes two key contributions. First, we perform a fully reproducible data processing pipeline to process images sourced from various museum collections. We annotate these images with carefully curated labels from biological databases and manual annotations to create an AI-ready dataset of visual traits, contributing to the advancement of AI in biodiversity science. Second, our proposed downstream tasks offer fertile grounds for novel computer vision research in addressing a variety of challenges such as long-tailed distributions, out-of-distribution generalization, learning with weak labels, explainable AI, and segmenting small objects. We benchmark the performance of several existing methods for our proposed tasks to expose future research opportunities in AI for biodiversity science problems involving visual traits.

\end{abstract}

%% file: sec/1_introduction.tex
\section{Introduction}
\label{sec:intro}

\begin{figure}
    \centering
    \includegraphics[width=\linewidth]{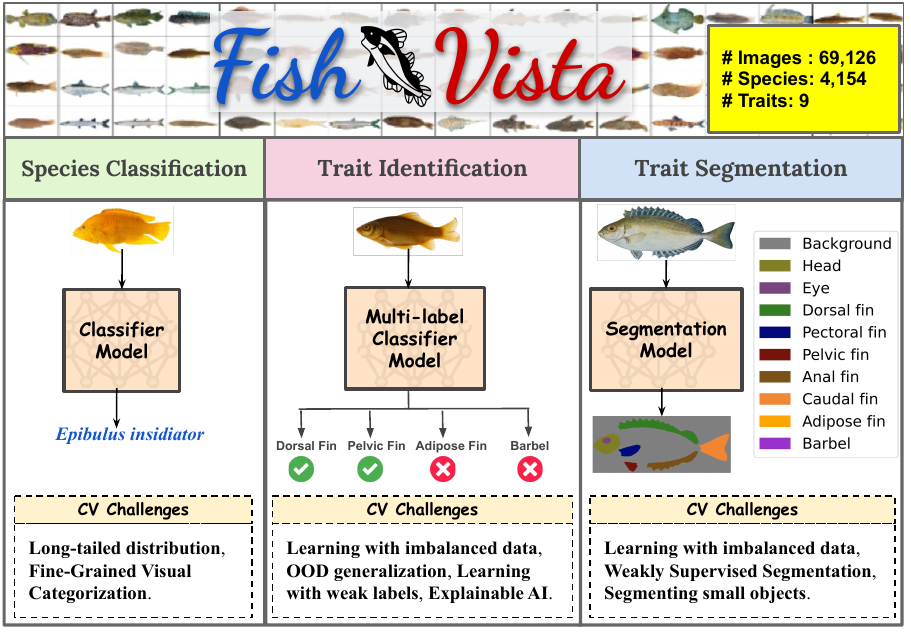}
    \caption{Overview of Fish-Vista tasks analyzing visual traits of fishes while exposing computer vision (CV) challenges.}
    \label{fig:teaser}
\end{figure}

In much the same way as large general-purpose image datasets such as ImageNet \cite{deng2009imagenet} have fueled the rise of deep learning in mainstream computer vision (CV) applications, the growing deluge of image datasets in organismal biology \cite{wah2011caltech, piosenka2023birds, stevens2024bioclip, gharaee2024bioscan,gharaee2024step,khosla2011novel,anantharajah2014local}
are poised to enable similar revolutions in the field of AI for biodiversity science \cite{tuia2022perspectives,fergus2024harnessing}. Images are increasingly being considered as the ``currency'' for documenting the vast
array of biodiverse organisms on our planet, with repositories containing millions of images of
biological specimens collected by scientists in field museums or captured by drones, camera traps, or
tourists posting photos on social media. This provides novel opportunities for CV research in biodiversity applications such as classifying species with fine-grained differences (termed Fine-grained Visual Categorization or FGVC \cite{zhang2022fine}), and segmenting the body of organisms in natural habitat images with complex backgrounds.

While these applications serve critical use-cases in biodiversity science, there are many other biologically relevant applications that have not been fully explored by current research in CV. For example, a grand challenge in biology is to discover characteristics of organisms, or \textit{traits} (e.g., {beak color, stripe pattern, and fin curvature}), that help in discriminating between species and understanding how organisms evolve and adapt to their environment  \citep{houlerossoni2022}. While some traits are behavioral, physiological, or related to the internal anatomy of organisms, we are interested in the analysis of traits that are externally visible in images, termed \textit{visual traits}. Detecting visual traits and localizing their presence from large collections of biodiversity images offers novel opportunities for CV research to advance our understanding of the impacts of climate change \cite{graham2009implications} on morphological features of organisms, and exploring the genetic and evolutionary underpinnings of their variations \cite{houle2022complexity}.

Current image datasets in biodiversity science suffer from two critical gaps that make them unfit for analyzing visual traits. \textit{First}, current biodiversity datasets predominantly focus on the task of species classification and do not include trait-level annotations. This is because while it is relatively easy to identify and document the species of an organism, obtaining annotations of visual traits (either at the image-level or pixel-level) requires expert knowledge and labor-intensive manual processing.
As a result, even though some datasets provide segmentation annotations of the entire body of organisms \cite{wah2011caltech, parkhi2012cats, saleh2020realistic} (which are easier to generate using modern annotation tools such as Segment Anything Model or SAM \cite{kirillov2023segment}), they do not provide annotations of visual traits that are smaller in size and more difficult to delineate, despite their importance in discovering biodiversity patterns.
\textit{Second}, while most datasets contain images of organisms captured in natural habitats \cite{saleh2020realistic, fishbase}, they generally do not contain images taken in controlled environments with uniform backgrounds, necessary for the analysis of fine-grained visual traits. In particular, the presence of complex backgrounds such as dense foliage or underwater elements in poor-lighting environments can occlude and obfuscate visual traits that are already hard to detect. Additionally, AI models trained on natural habitat images may learn to predict traits based on background patterns found in the habitats of certain species (e.g., presence of sky or ocean), introducing unintentional biases in the localization of visual traits within the body of organisms. 

To address these gaps, we introduce \textbf{Fish}-\textbf{Vis}ual \textbf{T}rait \textbf{A}nalysis (Fish-Vista), the first organismal dataset designed for the analysis of visual traits of fishes directly from images using novel problem formulations in CV. Fish-Vista contains 69,126 annotated images spanning 4,154 fish species, curated and organized to sever three downstream tasks of species classification, trait identification, and trait segmentation, as shown in Figure \ref{fig:teaser}. Our work makes two key contributions to the field of CV.
\textit{First}, we introduce a novel and fully reproducible data processing pipeline to convert images sourced from various museum collections including GLIN \cite{glin}, IDigBio \cite{IDigBio}, and MorphBank \cite{Morphbank} to create an ``AI-ready'' dataset of visual traits, a brand-new concept in AI for biodiversity science. \textit{Second}, our proposed downstream tasks offer fertile grounds for novel CV research in addressing a variety of challenges such as \textit{long-tailed distributions} (across all three tasks), \textit{out-of-distribution generalization} (trait identification), \textit{learning with weak labels} (trait identification), \textit{explainable AI} (trait identification), and \textit{segmenting small objects} (trait segmentation). We benchmark the performance of several existing methods for our proposed tasks to expose current gaps in CV research and to motivate future research directions in CV for answering biologically relevant questions important for the analysis of visual traits.

%% file: sec/2_related_works.tex
\section{Related Works}

\begin{table*}[ht]
\centering
\setlength\tabcolsep{5pt} 
\fontsize{8pt}{10}\selectfont
\renewcommand{\arraystretch}{0.9}
\resizebox{\textwidth}{!}
{
\begin{tabular}{lccccccc}
\toprule
\textbf{Dataset} & \textbf{Organism} & \textbf{\# Species} & \textbf{\# Images} & \parbox[t]{1.5cm}{\centering \textbf{Full-body\\Segmentation}} & \parbox[t]{1.5cm}{\centering \textbf{Visual Trait\\Information}} & \parbox[t]{1.5cm}{\centering \textbf{Visual Trait\\Segmentation}} & \textbf{Background} \\
\toprule
\textbf{CUB-200-2011} \cite{wah2011caltech} & Birds & 200 & 11,788 & $\checkmark$ & $\checkmark$ & x & Natural Habitat \\
\textbf{Birds 525} \cite{piosenka2023birds} & Birds & 525 & 89,885 & x & x & x & Natural Habitat \\
\textbf{NABirds} \cite{7298658} & Birds & 555 & 48,562 & x & $\checkmark$ & x & Natural Habitat \\
\textbf{Stanford dogs} \cite{khosla2011novel} & Dogs & 120 & 20,580 & x & x & x & Natural Habitat \\
\textbf{Oxford Pet} \cite{parkhi2012cats} & Cats, Dogs & 37 & 7,349 & $\checkmark$ & x & x & Natural Habitat \\
\textbf{FathomNet} \cite{katija2022fathomnet} & Marine Species & 2244 & 84,454 & x & x & x & Natural Habitat \\
\midrule
\textbf{Ulucan et al.} \cite{ulucan2020large} & Fish & 9 & 9,000 & $\checkmark$  & x & x & Controlled \\
\textbf{QUT Fish} \cite{anantharajah2014local} & Fish & 468 & 3,960 & x & x & x & Mixed \\
\textbf{DeepFish} \cite{saleh2020realistic} & Fish & NA & 39,766 & $\checkmark$ & x & x & Natural Habitat \\
\textbf{Fish4Knowledge} \cite{DBLP:series/isrl/104} & Fish & 23 & 27,370 & x & x & x & Natural Habitat \\
\textbf{FishBase} \cite{fishbase} & Fish & 35,600 & 64,000 & x & $\checkmark$ & x & Natural Habitat \\
\textbf{iNaturalist-2021-Fish} \cite{van2021benchmarking} & Fish & 183 & 46,996 & x & x & x & Natural Habitat \\ 
\textbf{FishNet} \cite{khan2023fishnet} & Fish & 17,357 & 94,778 & x & x & x & Natural Habitat \\ \midrule
\textbf{Fish-Vista (Ours)} & Fish & 4154 & 69,126 & $\checkmark$ & $\checkmark$ & $\checkmark$ & Controlled + Uniform \\
\bottomrule
\end{tabular}
}
\caption{Summary of commonly used fine-grained biodiversity datasets comprising images of organisms.}

\label{tab:datasets-comparision}
\end{table*}

Table \ref{tab:datasets-comparision} summarizes a number of biodiversity image datasets that have been published in the last two decades covering diverse categories of organisms such as birds, cats, dogs, and fishes, with varying numbers of images and species.
One of the common CV tasks explored with these datasets is to perform fine-grained visual categorization (FGVC) of species, which involves differentiating between closely related species based on subtle visual differences. 
However, as shown in Table \ref{tab:datasets-comparision}, most biodiversity datasets do not include trait-level information either at the level of pixels or images or species.
While some datasets such as CUB \cite{wah2011caltech}, Oxford Pets \cite{parkhi2012cats}, Ulucan et al. \cite{ulucan2020large}, and DeepFish \cite{saleh2020realistic} contain segmentation annotations of the whole body of organisms, they do not provide pixel-level annotations of individual traits that are fine-grained and smaller in size.
There are also datasets such as CUB \cite{wah2011caltech}, NABirds \cite{7298658}, and FishBase \cite{fishbase} that contain trait information at the level of images or species, but do not include trait segmentation annotations.
 A number of large-scale biodiversity datasets such as Tree-of-Life \cite{bioclip} and BioSCAN \cite{gharaee2024bioscan,gharaee2024step} have also been recently released containing several millions of images. However, they still lack annotations of visual traits at the level of pixels or images.


Another common feature in most biodiversity datasets is their focus on natural habitat images. For example, several fish image datasets feature fishes in their natural underwater habitats \cite{saleh2020realistic,DBLP:series/isrl/104,fishbase,van2021benchmarking}. While they are important for monitoring species populations out in the wild, they are not conducive to the analysis of visual traits of organisms because underwater images lack clarity and visual traits are obscured. 
While some datasets like QUT Fish \cite{anantharajah2014local} and Ulucan et al. \cite{ulucan2020large} feature fish images in controlled environments, they are limited in their number of images and diversity of fish species. 
A related work focusing on the analysis of visual traits of fishes includes FishShapes \cite{price2022fishshapes}, which provides numeric data of the lengths of various fish parts. However, it does not provide accompanying images of fishes used for making trait measurements. Another notable dataset in the domain of studying fish traits is FishBase \cite{fishbase}, which stands out as a comprehensive dataset comprising 64K images spanning 35K species. However, FishBase is limited in the number of images available per species poses a challenge for training modern AI models to analyze visual traits. Recently, a new dataset of FishNet was proposed in \cite{khan2023fishnet}, which combines fish images from the iNaturalist fish dataset \cite{van2021benchmarking} and FishBase annotations \cite{fishbase} with functional traits such as ecological and habitat information of species. However, these traits are not localizable in the images of organisms, and hence are outside our focus on visual traits. 
In contrast to all previous works, our proposed Fish-Vista dataset provides fine-grained trait annotations at species, image, and pixel levels, with controlled and uniform backgrounds from museum collections covering a large number of images across a wide range of species, as shown in Table \ref{tab:datasets-comparision}.

%% file: sec/3_methods.tex
\section{Fish-Vista Dataset}

\subsection{Why Fish-Vista?}
Fish-Vista fixes a critical gap in current benchmark datasets available in AI for Biodiversity Science by bridging high-quality images cleaned and curated from diverse museum collections with labels of visual traits obtained through expert annotations and knowledge-bases. Along with enabling a range of trait-related questions in the field of biodiversity science, a primary motivation behind Fish-Vista is to expose novel problem formulations and research challenges in CV tasks involving visual traits. For example, while there has been considerable work in FGVC for species classification, the connection between the subtle differences in species discovered by AI models and visual traits known to biologists has still not been established. We hope that by focusing on visual traits, our work advances the field of CV to focus on the explainability of fine-grained features that are localized in images and grounded in  knowledge of biological traits.

\subsection{Data Sources used in Fish-Vista}
We consider museum collections of fish images publicly available at \href{https://greatlakesinvasives.org/portal/index.php}{GLIN} \cite{glin, inhs, osum, fmnh, jfbm, ummz, umadison}, \href{http://www.idigbio.org/portal}{IDigBio} \cite{IDigBio}, and \href{https://www.morphbank.net/}{Morphbank} \cite{Morphbank} databases. We acquired these images along with their associated metadata including species names and licensing information from the \href{https://fishair.org/}{FishAIR} \cite{fishair} repository. In total, we collected 56,481 images from GLIN, 41,505 from IDigBio, and 9,000 from MorphBank. 


\begin{figure*}[ht]
  \centering
  \includegraphics[width=0.9\textwidth]{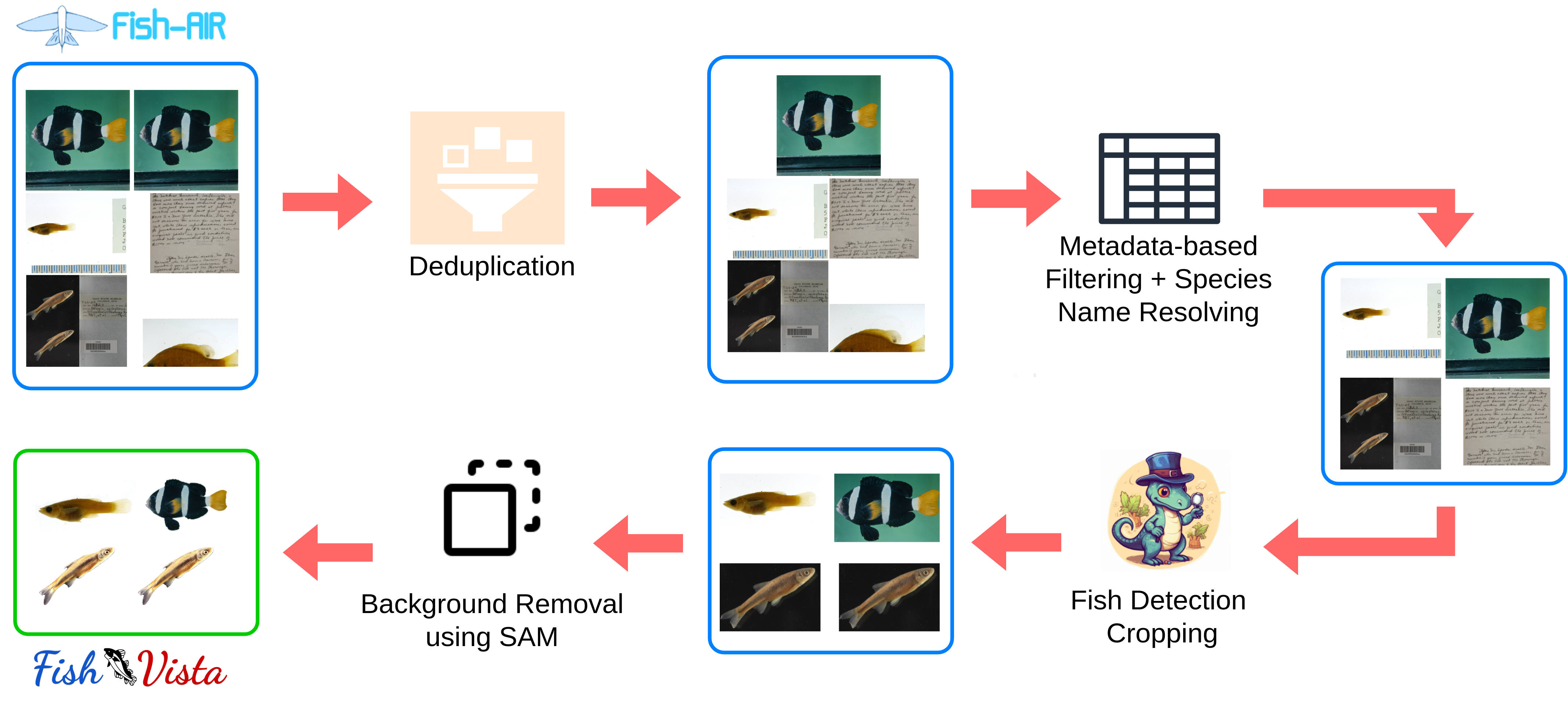}
  \caption{An overview of the data processing pipeline used to create Fish-Vista.}
  \label{fig:processing-pipeline}
\end{figure*}
\subsection{Data Processing Pipeline}
\label{section:data-processing}


There are two key challenges with FishAIR images that we need to address: (1) museum images contain several visual elements such as rulers and tags apart from fish specimens that need to be cropped, and (2) there are many noisy images in museum collections including hand-written notes and radiographic images that need to be dropped. Figure \ref{fig:processing-pipeline} shows a schematic of our processing pipeline to address these challenges comprising of the following five steps. 

\par \noindent \emph{1. Removing Duplicates:}
Since museum collections sometimes contain duplicate images stored under different filenames, we remove duplicate images with same MD5-checksum, to avoid data leakage in training and test splits. 


\par \noindent \emph{2. Quality Metadata-based Filtering:} For a subset of the raw images ($\approx 30k$), we obtained manually annotated \textit{quality metadata} from FishAIR that includes information about the visibility of all parts of a specimen and the orientation of the fish (e.g., side-view or top-view). We use this data to filter images where all visual traits are not visible.



\par \noindent \emph{3. Filtering Noisy Species Names:} Scientific species names of FishAIR images sometimes contain inaccuracies like typographical errors or synonymous names. To mitigate this, we exclude entries with species names that are not valid strings, such as \textit{gen. sp.}. We utilize the Open Tree Taxonomy (OTT) \cite{opentreeoflife2019} to correct typographical errors and standardize synonyms to their canonical forms, ensuring consistent categorization of species names. 


\par \noindent \emph{4. Detecting and Cropping Fish Bounding Boxes:} 
We use Grounding DINO \cite{liu2023grounding}, a SOTA zero-shot object detection model, to detect and crop  tight bounding boxes around fish specimens  in the museum images. 
We discard bounding boxes with either dimensions smaller than 224 pixels to avoid low-resolution images.



\par \noindent \emph{5. Removing Background using SAM:} The backgrounds of fish bounding boxes often contain features unique to specific species or museum collections, introducing biases in the data that are not useful for analyzing visual traits. 
To avoid this, we use the Segment Anything Model (SAM) \cite{kirillov2023segment} to segment the whole body of a fish specimen from its bounding box and use a uniform white background. 

Further details of the processing pipeline along with quantitative and qualitative validations for \textit{Step 4} and \textit{Step 5} are provided in the Appendix.
We finally obtain 100K images spanning 10K species that we use in the three downstream tasks as described in the following. 


\subsection{Fish-Vista Tasks}

\begin{figure}[ht]
    \centering
    \includegraphics[width=0.48\textwidth]{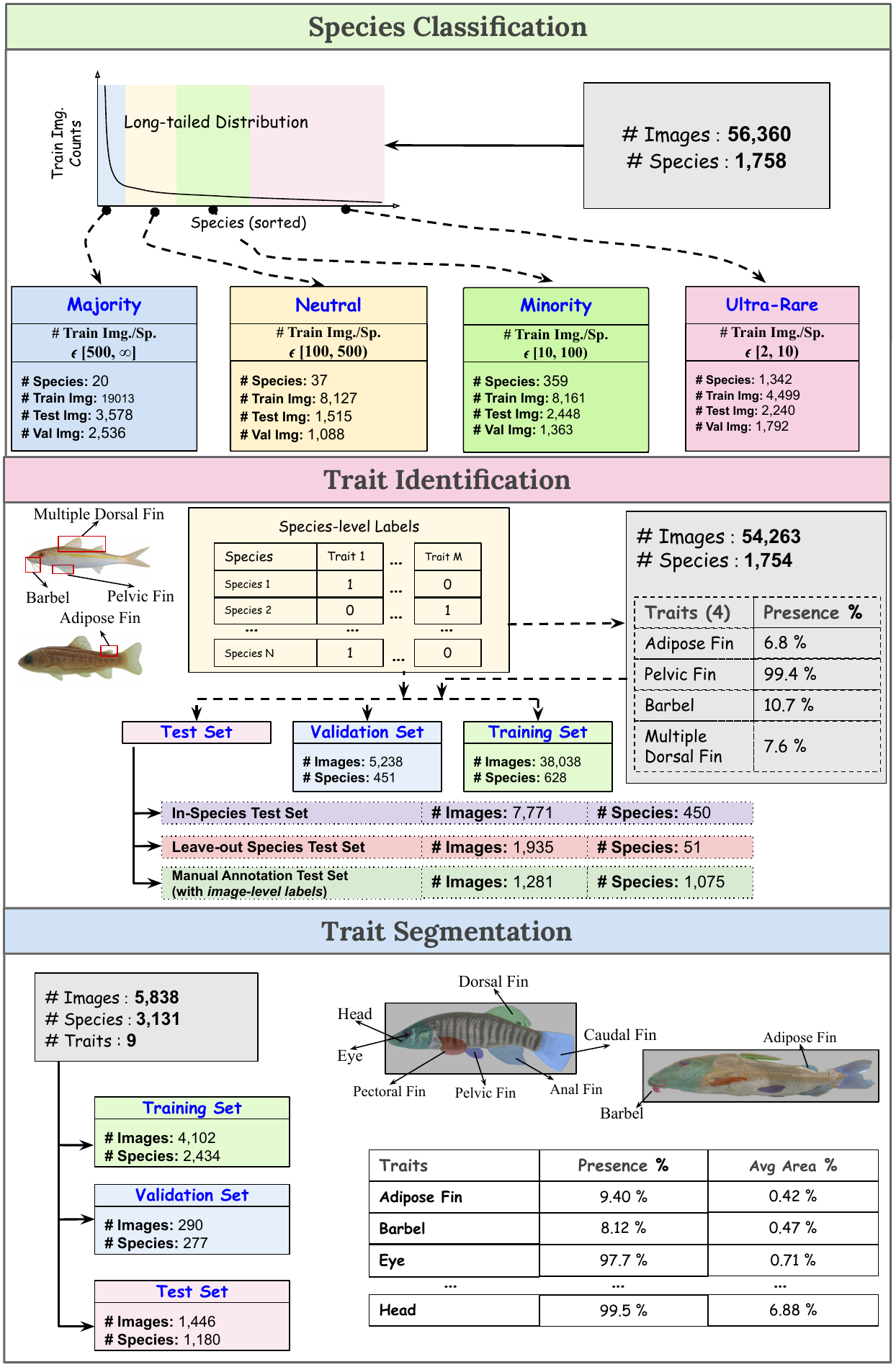}
    \caption{An overview of the key statistics of Fish-Vista Dataset.}
    \label{fig:key_stat}
\end{figure}

Figure \ref{fig:key_stat} shows an overview of the process followed for creating data partitions in the three Fish-Vista tasks exploring different CV research challenges in each task, along with key statistics.

\subsubsection{Fine-grained Species Classification}
Species classification involves categorizing images to their respective species by distinguishing fine-grained visual traits (Figure \ref{fig:teaser}, Species Classification). One of the key challenges in species classification with Fish-Vista is the extreme long-tailed nature of image count distributions across all species classes. We perform a sequence of operations to prepare relevant partitions of the Fish-Vista dataset for training and evaluating AI models for species classification while accounting for the long-tailed distribution. We first remove species that have less than 4 images per class, to ensure sufficient number of images for training, testing, and validation. The remaining species still suffered from  a high degree of class imbalance as some species contain up to 2K images while many others have less than 10 images. To further remove rare species that do not have representative high-quality images, we manually inspect the visual quality of a randomly sampled subset of 15\% images for each species. Species with predominantly low-quality images in these subsamples or those lacking clear visual traits are dropped from the analysis (see Appendix for additional details of this filtering step). This results in the final \emph{FV-Classification} dataset, which contains 56,360 images spanning 1,785 species. 

We construct train, validation, and test splits using stratified sampling across every species using splitting fractions of 75\%, 15\%, and 10\% for train, test, and val sets, respectively. We manually inspected every image in the test set to ensure that they are of high quality. To address the dataset's highly imbalanced long-tailed distribution, we categorize the 1,785 species in FV-Classification into four groups based on their training image counts per species: \textit{Majority} (500 or more images), \textit{Neutral} (100-499 images), \textit{Minority} (10-99 images), and \textit{Ultra-Rare} (fewer than 10 images). Figure \ref{fig:key_stat} provides statistics on these four data splits of FV-Classification and their distributions of training, test, and validation images. Note that while we only have 20 majority species, we have 1,342  Ultra-rare species, demonstrating the highly imbalanced long-tailed nature of FV-Classification. Additional details about the creation of data splits, manual test set filtering, and data distributions are provided in the Appendix. 

\textbf{Key CV challenges:} Given the large number of fish species that have varying evolutionary and anatomical similarities, classification models must differentiate subtle, fine-grained visual differences among highly similar fish species, making this a challenging fine-grained visual categorization task. The dataset’s long-tailed distribution adds another layer of complexity to the problem of species classification.

\subsubsection{Trait Identification}
Trait identification is the task of predicting the presence or absence of visual traits from an image (Figure \ref{fig:teaser}, Trait Identification). There are three key points to consider for trait identification. \textit{First}, predicting presence/absence of all possible traits in fish images is unnecessary; we only need to focus on traits that vary across species and are considered biologically \emph{interesting}. Traits deemed \emph{interesting} are often ``rare" traits -- those that are present or absent in only a few species. For example, the presence of eyes, which is universal across all fish species, is neither informative nor biologically significant to predict, whereas rare traits, such as the presence of an adipose fin (see Figure \ref{fig:key_stat} for definition), offers more scientific value. \textit{Second}, manually annotating thousands of images for trait presence/absence requires biological expertise, is time-consuming and difficult to scale. \textit{Third}, the effectiveness of trait identification models should be evaluated on out-of-distribution sets containing species not seen during training. This would ensure that our models learn to generalize based on the visual appearance of traits, rather than memorizing species names and predicting traits known to be associated with every species. Furthermore, we should also evaluate the ability of models to accurately \emph{identify} or visually \emph{localize} the traits within the image while predicting their presence or absence.

We create the trait identification dataset with the aforementioned key points in mind involving a number of steps as outlined in Figure \ref{fig:key_stat}. \textit{First}, we select four scientifically significant traits that vary across species -- adipose fin, pelvic fin, barbel, and multiple dorsal fins (Figure \ref{fig:key_stat}). We can see that some of these traits are present over a large percentage of images (e.g., pelvic fin), there are traits that are rare such as adipose fin and multiple dorsal fin that are present only 6.8\% and 7.6\%, respectively. \textit{Second}, instead of annotating each image manually, we gather species-level trait labels from the Phenoscape KnowledgeBase (KB) \cite{mabee2018phenoscape} and FishBase \cite{fishbase}. We obtain information for 682 species and map this information to our images. Since Fish-Vista images have been processed to include complete fish specimens with all visible traits, species-level labels provide a reasonable basis for image-level trait identification. This species-level labeling introduces an additional layer of complexity, requiring models to identify traits at the image level based on coarse-grained, species-level annotations. After following these steps, the identification dataset, FV-Id, contains 53K images across 682 species. 

We split FV-Id dataset to create train/test/val splits. We ensure that the test set obtained at this stage entirely consists of species encountered during training, and we refer to this as the \textit{in-species} test set. Additionally, to further evaluate the generalization performance of trait identification across unseen species, we create two extra leave-out-species test sets -- \textit{leave-out-species} and \textit{test-manual-annotation}. The \textit{leave-out-species} dataset is created by holding out 51 species (1,935 images) from the identification dataset that we do not use for training. The \textit{manual-annotation} dataset consists of 1,281 manually annotated images across 1,075 species, labeled by expert biologists for the presence or absence of the four target traits. We choose a widely diverse set of species that are not seen during training to evaluate the generalization power of trait identification models. We also include manual pixel-level segmentation annotations for the \textit{manual-annotation} set to enable evaluation of trait localization performance within the body of fish images. Adding the \textit{manual-annotation} set results in the final identification dataset results in a total of 54,263 images across 1,754 species. Key statistics for the dataset across the various data splits and four trait labels are provided in Figure \ref{fig:key_stat}.  Additional details in the creation of FV-Id are provided in the Appendix. 

\textbf{Key CV Challenges:} We test the generalization performance of models not just on species seen during training (representing in-distribution performance) but also new species never seen before during training (representing out-of-distribution performance). We also have high degree of imbalance in the presence\% of some traits such as adipose fin and multiple dorsal fin. By training AI models using species-level trait labels and evaluating their performance on a test set with manual image-level annotations, FV-Id is enabling the study of learning with weak (coarse-scale) labels.

\subsubsection{Trait Segmentation}
Going beyond trait identification at the image level, we introduce the task of trait segmentation, where the goal is to precisely localize and delineate traits within the fish images. We focus on segmenting nine visible traits on fish bodies that are well-suited for applying CV models, while also carrying significance for the scientific analysis of traits. See Figure \ref{fig:teaser} and Figure \ref{fig:key_stat}. It is worth noting that while certain traits, such as the \textit{eye}, may be uninformative for image-level presence/absence prediction in trait identification, their localization on images is still significant.
We obtain the segmentation dataset (FV-Segmentation) by manually labeling the 9 traits across 5,838 images. The annotation process is conducted by expert biologists by utilizing the CVAT tool \cite{cvat2023}. Due to the labor-intensive nature of the annotation process and the need for biological expertise, the segmentation dataset is smaller than its classification and identification counterparts. To enhance segmentation models' ability to generalize from this limited dataset, we include a highly diverse set of 3,131 species in the annotations. We provide additional details in the Appendix. 

\textbf{Key CV Challenges:} The trait segmentation task presents several unique challenges. First, because of the relatively smaller size of the dataset, the segmentation models must be adept at learning to generalize using limited number of labels. Second, the various fins of the fish can appear visually similar in shape and texture. This means models must rely on positional cues alone to distinguish between these traits, which can cause misclassifications. This is further complicated by anatomical variations across diverse species. For example, the adipose and dorsal fins can look similar, and also appear in similar positions across various species. Third, some traits are very small, posing the well-known challenge of small object segmentation. For example, the barbel is a whisker-like organ near the fish's head and occupies a small number of pixels. Finally, as with identification, the presence of certain traits is very rare, creating high imbalance for those traits. For example, adipose fin and barbel, which are both small traits, are also very rare, combining the challenges of small object segmentation with data imbalance.

%% file: sec/4_results.tex
\begin{figure}[h]
\centering
\begin{minipage}{0.48\textwidth}
    \includegraphics[width=0.9\textwidth]{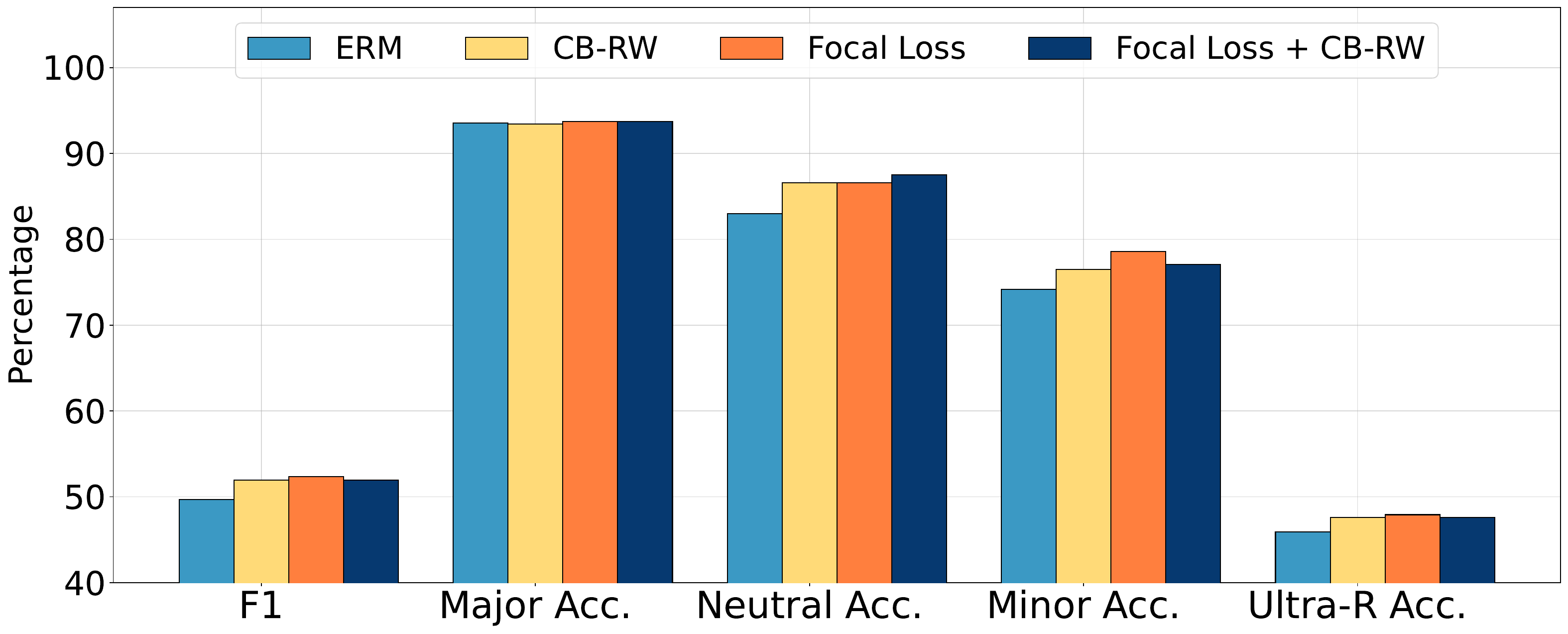}
        \captionof{figure}{\small{Comparison of the fine-grained classification performance of different imbalanced classification methods.}}
    \label{fig:cls_imbalance}
    \vfill
    \vspace{2ex}
    \centering
\setlength\tabcolsep{0.08cm} 
\renewcommand{\arraystretch}{0.9}
\resizebox{0.8\textwidth}{!}
{    
\begin{tabular}{lccccc}
    \toprule
        
        \fontsize{9.2pt}{10}\selectfont\textbf{Model} & \fontsize{9pt}{10}\selectfont\textbf{F1} & \begin{tabular}[c]{@{}c@{}}{\fontsize{9pt}{10}\selectfont\textbf{Major}} \\
        {\fontsize{9pt}{10}\selectfont\textbf{Acc.}}\end{tabular} & \begin{tabular}[c]{@{}c@{}}{\fontsize{9pt}{10}\selectfont\textbf{Neutral}} \\ {\fontsize{9pt}{10}\selectfont\textbf{Acc.}}\end{tabular} & \begin{tabular}[c]{@{}c@{}}{\fontsize{9pt}{10}\selectfont\textbf{Minor}} \\ {\fontsize{9pt}{10}\selectfont\textbf{Acc.}}\end{tabular} & \begin{tabular}[c]{@{}c@{}}{\fontsize{9pt}{10}\selectfont\textbf{Ultra-R}} \\
        {\fontsize{9pt}{10}\selectfont\textbf{Acc.}} \end{tabular}\\ \midrule
        \fontsize{8.5pt}{10}\textbf{VGG-19} \cite{simonyan2014very} & 49.7 & 93.5 & 83.0 & 74.2 & 45.9 \\
        \fontsize{8.5pt}{10}\textbf{ResNeXt-50} \cite{xie2017aggregated} & 44.4 & 91.4 & 78.3 & 69.8 & 39.1 \\ 
        \fontsize{8.5pt}{10}\textbf{RegNetY-4G} \cite{radosavovic2020designing} & 43.7 & 89.8 & 77.4 & 68.5 & 38.5 \\ 
        \midrule
        \fontsize{8.5pt}{10}\textbf{ViT-B16} \cite{ViT} & 48.3 & 88.7 & 82.3 & 73.3 & 43.4 \\
        \fontsize{8.5pt}{10}\textbf{Swin-B-22k} \cite{swin} & 55.1 & 92.6 & 86.2 & \colorbox{blue!15}{79.6} & \colorbox{blue!15}{50.4} \\
        \fontsize{8.5pt}{10}\textbf{CvT-13} \cite{cvt} & 49.3 & 92.0 & 83.3 & 73.5 & 44.7 \\
        \fontsize{8.5pt}{10}\textbf{MaxViT-T} \cite{maxvit} & \colorbox{blue!35}{57.8} & \colorbox{blue!15}{94.4} & \colorbox{blue!35}{86.7} & \colorbox{blue!35}{81.4} & \colorbox{blue!35}{53.9} \\
        \fontsize{8.5pt}{10}\textbf{PVT-v2-b0} \cite{wang2022pvt} & 51.0 & 92.0 & 83.4 & 75.7 & 45.8 \\
        \midrule
        \fontsize{8.5pt}{10}\textbf{BioCLIP-ZS} \cite{bioclip} & 4.6 & 1.1 & 1.4 & 10.2 & 5.6 \\
        \fontsize{8.5pt}{10}\textbf{CLIP-ZS} \cite{clip} & 0.1 & 0.0 & 0.3 & 0.4 & 0.2 \\
        \midrule
        
        \fontsize{8.5pt}{10}\textbf{BioCLIP-LP} \cite{bioclip} & 38.2 & \colorbox{red!15}{75.5} & \colorbox{red!15}{65.2} & 61.3 & 31.1 \\
        
        \fontsize{8.5pt}{10}\textbf{CLIP-LP} \cite{clip} & \colorbox{red!15}{25.4} & \colorbox{red!35}{55.9} & \colorbox{red!35}{49.8} & \colorbox{red!15}{46.7} & \colorbox{red!15}{20.9} \\

        \fontsize{8.5pt}{10}\textbf{DINOv2-LP} \cite{oquab2023dinov2} & 53.1 & 89.9 & 78.04 & 76.04 & 47.02 \\
        
        \midrule
        \fontsize{8.5pt}{10}\textbf{INTR} \cite{paul2023simple} & \colorbox{red!35}{6.1} & 92.2 & 73.2 & \colorbox{red!35}{22.6} & \colorbox{red!35}{0.62} \\
        \fontsize{8.5pt}{10}\textbf{TransFG} \cite{he2022transfg} & 50.3 & \colorbox{blue!35}{94.5} & \colorbox{blue!15}{86.6} & 75.5 & 45.3 \\
        \bottomrule
    \end{tabular}
}
\captionof{table}{\small{Comparison of classification performance  (in \%). Results  are color-coded as \colorbox{blue!35}{Best}, \colorbox{blue!15}{Second best}, \colorbox{red!35}{Worst}, \colorbox{red!15}{Second worst} (excluding Zero-Shot methods \textbf{(ZS)}).}}
\label{tab:classification}
\end{minipage}
\end{figure}

\section{Experiments and Results}

We compare results of SOTA methods on the three Fish-Vista tasks in the following. Our goal is not to determine the best-performing model for each task, but rather to discover insights and highlight key challenges that current CV methods encounter in Fish-Vista. 

\subsection{Species Classification}
We evaluate performance on this task using a wide range of approaches including CNN-based and Transformer-based backbones, fine-grained categorization methods, vision foundation models, and class-imbalance techniques. We report the overall macro-averaged F1-score and mean class accuracy for each subcategory to assess performance across the imbalanced distribution. Key results are shown in Table \ref{tab:classification}, with additional results in the Appendix.
Most methods achieve around 50\% F1 on the test set. As expected, most methods perform well on \textit{Majority} species $(\approx 90\%)$ and \textit{Neutral} species $(\approx 80\%)$, but the performance drops significantly on rare categories, with \textit{Ultra-rare} species reaching only about 50\% accuracy.

We also evaluated the zero-shot performance of CLIP \cite{clip} and BioCLIP \cite{bioclip}, a foundation model for biological  images. Both models performed near random on Fish-Vista, with BioCLIP outperforming CLIP, particularly on \textit{Rare} species, still with low accuracy. Next, we use pre-trained features from CLIP, BioCLIP, and Dino-v2 to perform linear probing by training a single classification layer. This is where we observe the foundational effect of BioCLIP features, which significantly outperforms CLIP. Linear probing with Dino-v2 performs better than both CLIP and BioCLIP. However, it still performs worse than the best-performing backbone models. These results indicate that Fish-Vista presents new opportunities for improving vision foundation models in biology.

\begin{figure}[ht]
\centering
\begin{minipage}{0.48\textwidth}
    \centering
    \includegraphics[width=0.9\textwidth]{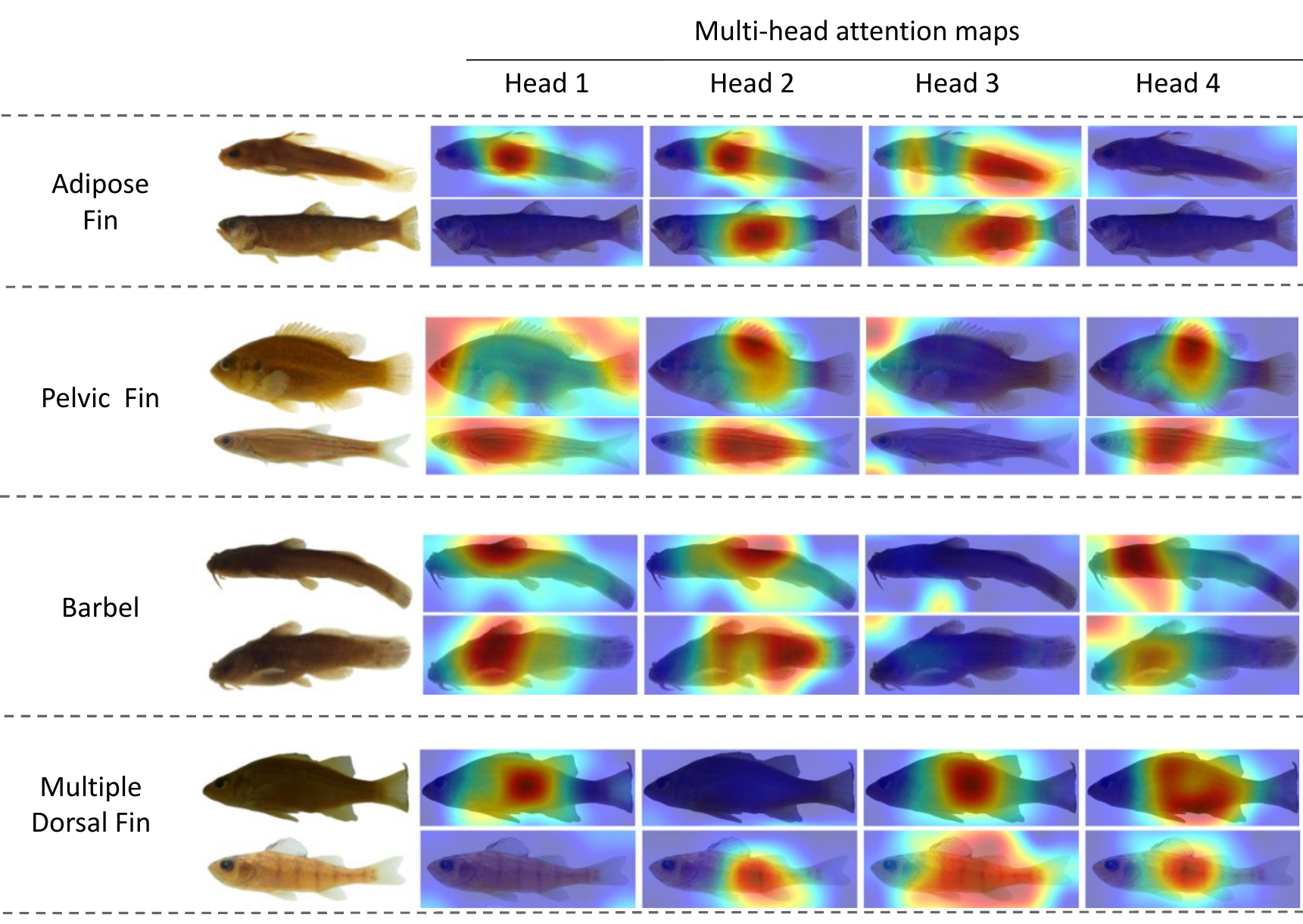}
        \captionof{figure}{\small{Attention maps from the Query2Label-SWIN model corresponding to the four traits.}}
    \label{fig:attention-q2l-swin}
    \vfill
    \vspace{2ex}
    \centering
\setlength\tabcolsep{0.08cm} 
\renewcommand{\arraystretch}{0.9}
\resizebox{\textwidth}{!}
{    
\begin{tabular}{lcccccc}
\toprule
\multirow{2}{*}{\renewcommand{\arraystretch}{1.2}{\begin{tabular}[c]{@{}c@{}}\fontsize{8.5pt}{10}\selectfont\textbf{Q2L} \\ \fontsize{8.5pt}{10}\selectfont\textbf{Backbone}\end{tabular}}} & \multirow{2}{*}{\renewcommand{\arraystretch}{1.2}{\begin{tabular}[c]{@{}c@{}}{\fontsize{8.5pt}{10}\selectfont\textbf{ \# Attention}} \\ {\fontsize{8.5pt}{10}\selectfont\textbf{Heads}}\end{tabular}}} & \multicolumn{4}{c}{\fontsize{8.5pt}{10}\selectfont\textbf{IoU ($\times 100$)}}                    & \multirow{2}{*}{\renewcommand{\arraystretch}{1.2}{\begin{tabular}[c]{@{}c@{}}{\fontsize{8.5pt}{10}\selectfont\textbf{mIOU}} \\ {\textbf{($\times 100$)}}\end{tabular}}} \\ \cmidrule{3-6}
                              &                                   & \fontsize{8.5pt}{10}\selectfont\textit{Adipose} & \fontsize{8.5pt}{10}\selectfont\textit{Pelvic} & \fontsize{8.5pt}{10}\selectfont\textit{Barbel} & \fontsize{8.5pt}{10}\selectfont\textit{Dorsal} &                       \\ \midrule
\fontsize{8pt}{10}\textbf{ResNet}                        & 1                   & $0.014$   & $1.482$   & $0.0008$ & $1.731$   & $0.81$                  \\
\fontsize{8pt}{10}\textbf{ResNet}                        & 4                   & $0.003$   & $0.963$   & $0.0067$  & $0.908$   & $0.47$                 \\
\fontsize{8pt}{10}\textbf{SWIN-B}                     & 1                      & $0.007$    & $1.447$   & $0.005$  & $1.945$   & $0.85$                  \\
\fontsize{8pt}{10}\textbf{SWIN-B}                     & 4                      & $0.048$   & $1.844$   & $0.002$  & $0.967$   & $0.72$                 \\ 
\bottomrule
\end{tabular}
}
\captionof{table}{\small{IoU of Query2Label (Q2L) attention maps for each trait. The IoUs and mIoUs are amplified by 100 times. }}
\label{tab:attention-miou}
\end{minipage}
\end{figure}

Given the dataset’s highly imbalanced nature, we evaluated the impact of well-known imbalance-handling techniques, including class-balanced re-weighting (CB-RW) \cite{cui2019class} and focal loss \cite{lin2017focal}. Results using one of our top-performing backbones, VGG-19, are shown in Figure \ref{fig:cls_imbalance}. We observe that both techniques slightly improve performance on the neutral, minority and ultra-rare classes, while not hurting the majority performance. The results highlight the long-tailed challenge in Fish-Vista classification.

\textbf{Insights:} Standard classification techniques, including fine-grained and class-imbalance methods, may not perform optimally on Fish-Vista -- especially for the rare species that constitute the majority of the dataset -- due to its challenging long-tailed distribution and fine-grained categorization requirements.
We provide additional experiments and implementation details in the Appendix.

\begin{figure}
    \centering
    \includegraphics[width=0.4\textwidth, height=4.6cm]{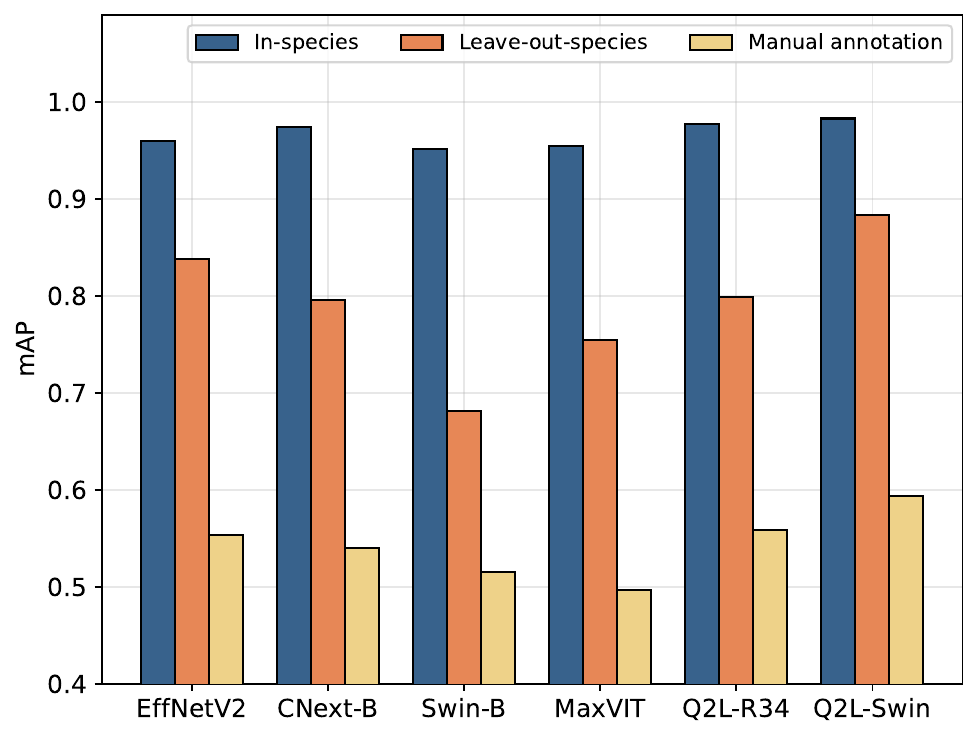}
    \caption{\small{Trait identification performance of different multi-label classification methods.}}
    \label{fig:identification_results}
\end{figure}

\begin{table*}[!htbp]
\small
\centering
\setlength\tabcolsep{1.5pt}
\renewcommand{\arraystretch}{1}
\resizebox{0.80\textwidth}{!}{
\begin{tabular}{lccccccccccc}
\toprule
\multirow{2}{*}{}{}  & \multirow{3}{*}{} & \multicolumn{10}{c}{\fontsize{8.5pt}{10}\textbf{Trait-wise IoU}} \\
\cmidrule(lr){3-12}
\renewcommand{\arraystretch}{0.2}{\begin{tabular}[l]{@{}l@{}} \fontsize{8.5pt}{10}\selectfont \textbf{Model} \\ {}\end{tabular}} & \renewcommand{\arraystretch}{0.2}{\begin{tabular}[l]{@{}l@{}} \fontsize{8.5pt}{10} \textbf{mIoU} \\ {}\end{tabular}} & \textit{BG} & \textit{Head}  & \textit{Eye} & \textit{Dorsal} & \textit{Pectoral} & \textit{Pelvic} & \textit{Anal} & \textit{Caudal} & \textit{Adipose} & \textit{Barbel}
    \\
    \midrule
        \fontsize{8pt}{10}\selectfont\textbf{PSPNet} \cite{zhao2017pspnet} & \colorbox{red!35}{70.5} & \colorbox{red!35}{93.6} & \colorbox{red!35}{80.4} & \colorbox{red!35}{75.9} & \colorbox{red!35}{82.4} & \colorbox{red!35}{61.7} & 78.6 & \colorbox{red!35}{79.7} & \colorbox{red!35}{86.6} & \colorbox{red!35}{49.1} & 16.8 \\ 
        \fontsize{8pt}{10}\selectfont \textbf{DeepLabV3} \cite{chen2017rethinking}& 73.9 & 94.5 & 83.2 & 76.6 & 85.1 & 67.6 & 80.7 & 83.6 & 87.8 & 59.4 & 20.2 \\ 
        \fontsize{8pt}{10}\selectfont \textbf{DeepLabV3Plus} \cite{deeplabv3plus2018} & 74.3 & 94.7 & 83.2 & 77.3 & 85.9 & 66.9 & \colorbox{blue!15}{82.6} & 84.3 & 88.7 & 58.4 & 20.8 \\ 
        \fontsize{8pt}{10}\selectfont \textbf{UNet} \cite{ronneberger2015u}& 74.4 & 95.0 & 83.6 & 78.3 & 86.7 & 67.2 & \colorbox{blue!35}{83.3} & \colorbox{blue!15}{84.8} & \colorbox{blue!15}{89.4} & 55.4 & 20.5 \\ 
        \fontsize{8pt}{10}\selectfont \textbf{Semantic FPN} \cite{kirillov2019panoptic}& 74.8 & 94.7 & 83.2 & 77.4 & 86.4 & 66.6 & \colorbox{blue!15}{82.6} & 84.6 & 88.6 & \colorbox{blue!15}{61.0} & \colorbox{blue!15}{22.4} \\ \hdashline
        \fontsize{8pt}{10}\selectfont \textbf{Mask2Former} \cite{cheng2021per}& \colorbox{blue!35}{86.6} & \colorbox{blue!35}{97.0} & \colorbox{blue!35}{90.2} & \colorbox{blue!35}{87.5} & \colorbox{blue!35}{91.9} & \colorbox{blue!35}{80.3} & 82.5 & \colorbox{blue!35}{88.4} & \colorbox{blue!35}{92.7} & \colorbox{blue!35}{78.0} & \colorbox{red!35}{0.0} \\ 
        \fontsize{8pt}{10}\selectfont \textbf{YOLOv8} \cite{varghese2024yolov8} & \colorbox{blue!15}{81.5} & \colorbox{blue!15}{96.4} & \colorbox{blue!15}{83.9} & \colorbox{blue!15}{81.5} & \colorbox{blue!15}{87.3} & \colorbox{blue!15}{76.7} & \colorbox{red!35}{77.1} & 84.3 & 87.8 & 60.1 & \colorbox{blue!35}{28.0} \\ \hdashline
        \fontsize{8pt}{10}\selectfont \textbf{MOLMO + SAM} \cite{deitke2024molmo, ravi2024sam} & 36.8 & 85.6 & 32.3 & 29.6 & 48.5 & 30.3 & 37.6 & 27.7 & 74.6 & 0.6 & 0.7 \\ \midrule
    \end{tabular}
    }
    \caption{\small{Performance (in \%) of seven mainstream segmentation models on the Segmentation dataset, along with a zero-shot architecture. Results are color-coded as \colorbox{blue!35}{Best}, \colorbox{blue!15}{Second best}, \& \colorbox{red!35}{Worst} (excluding Zero-Shot method \textbf{MOLMO + SAM}). }}
\label{tab:segmentation}
\end{table*}

\subsection{Trait Identification}

We train standard backbone models using a multi-label classification objective to predict the presence or absence of the four traits in the FV-Identification (FV-Id) dataset. Each model is evaluated on three FV-Id test sets: \textit{in-species}, \textit{leave-out-species}, and \textit{manual-annotated} (introduced previously). Figure \ref{fig:identification_results} compares the Mean Average Precision (mAP) of top-performing models across all test sets. Trait-wise results and additional metrics for an extensive range of models are provided in the Appendix.

Our best-performing models achieve high accuracy on the \textit{in-species} test set, as expected, given that these species were included in training. However, performance drops substantially on the \textit{leave-out-species} test set, and this decline is even more pronounced on the highly diverse \textit{manual-annotated} test set. These results indicate that existing methods struggle to generalize to unseen species, which is a key requirement for trait identification.

A critical aspect of the identification task is determining whether models can visually \textit{attend} to the correct traits—i.e., can we achieve trait localization through image-level trait identification? To assess this, we examine our top-performing identification model, Query2Label with SWIN backbone, using the model's transformer attention maps for each trait. We calculate the Intersection over Union (IoU) between ground-truth trait segmentations and the model's attention maps (Table \ref{tab:attention-miou}) and visualize the attention maps for images where the model correctly predicts traits (Figure \ref{fig:attention-q2l-swin}). Despite high accuracy, Query2Label demonstrates extremely low mIoUs and scattered attention maps, indicating a failure to attend to the correct traits. This highlights a limitation in current models: they may predict traits with decent performance at the image level but lack interpretability and spatial awareness necessary for localization. 

\textbf{Insights: } Existing methods struggle to generalize to unseen species and lack the ability to focus on relevant image regions -- an essential feature for model interpretability and visual grounding.

\subsection{Trait Segmentation}
We evaluate several mainstream image segmentation models for this task, including semantic segmentation architectures, instance segmentation models, and a zero-shot segmentation method. Table \ref{tab:segmentation} presents the overall mIoU and individual trait-wise IoUs for each method. Traits that have higher presence and occupy larger areas, such as the dorsal and caudal fins, generally achieve high IoUs of around 80\%. However, performance drops significantly for smaller, rarer traits like the adipose fin and barbel, and for traits that are located over the body, such as the pectoral fin.

All methods struggle particularly with the adipose fin and barbel. Notably, the best-performing model, Mask2Former, entirely fails  to detect the barbel. This difficulty is likely due to both traits being rare (low presence) and occupying minimal area (see Figure \ref{fig:key_stat}). Further inspection of the confusion matrix (Appendix) reveals that the barbel, located beneath the head, is frequently misclassified as the head, while the adipose fin, often near the dorsal fin, is misclassified as the dorsal fin. These results underscore the challenges that current methods face in accurately segmenting small, rare, and fine-grained traits in Fish-Vista.

Finally, we investigate the zero-shot segmentation capabilities of the Segment Anything Model (SAM-v2) \cite{ravi2024sam}, which relies on spatial prompts (e.g., points) for segmentation. Using the large vision-language model (LVLM) Molmo \cite{deitke2024molmo}, we generate these spatial prompts by directing Molmo to identify trait locations in images through textual prompts. The points generated by Molmo then serve as input for SAM-v2 to segment the traits. While we did not expect high performance, results demonstrate promising mIoU on traits like the \textit{dorsal} and \textit{caudal fins}. 

\textbf{Insights: } Conventional segmentation methods face significant challenges in localizing small, rare and fine-grained traits in Fish-Vista. Moreover, large foundational models like LVLMs and SAM have the potential to localize scientifically relevant visual traits.

%% file: sec/5_conclusion.tex
\section{Limitations of Fish-Vista}
While the processing pipeline of Fish-Vista includes a range of filtering steps, there may still be some images that are noisy and do not clearly exhibit visual traits, such as those with deformed fins (see Appendix for example images). Additionally, in the task of trait identification, while we assume that species-level labels of the presence or absence of traits are representative over all images of the species, this may not be true especially when certain traits in an image specimen are occluded due to poor data quality. We show some of these examples in the Appendix. Finally, our results on the Trait Segmentation task are limited by the relatively smaller number of labeled images compared to the other two tasks, which is due to the labor-intensive nature of generating pixel-level trait segmentation annotations.



\subsection{Future Research Directions}
We provide hierarchical taxonomic information for most of our species, similar to iNaturalist and FishNet. We preserve the raw URL for all images in Fish-Vista, allowing users access to the original images (with background and surroundings). Besides visual trait analysis, Fish-Vista offers potential for many additional applications. For example, the dataset can enable the integration of taxonomic information into CV models, similar to PhyloNN \cite{elhamod2023discovering}. Additionally, Fish-Vista can serve as a valuable resource to train foundation models for biology, similar to BioCLIP \cite{bioclip}. 



\section{Acknowledgments}
This research is supported by grants from the National Science Foundation (NSF) for the HDR Imageomics Institute (OAC-2118240). We are thankful for the support of computational resources provided by the Advanced Research Computing (ARC) Center at Virginia Tech and the Ohio Supercomputer Center.

%% file: sec/X_suppl.tex
\clearpage
\setcounter{page}{1}
\maketitlesupplementary

\appendix

\section*{Appendix}

\section{Code and Dataset}
We provide the experiment code, metadata files for all splits of Fish-Vista, and some example images and annotations in the following anonymous github link: \href{https://anonymous.4open.science/r/fish-vista-anonymized-19C8}{https://anonymous.4open.science/r/fish-vista-anonymized-19C8} . Due to the memory limitations of GitHub and the constraints of supplementary submission requirements, we are unable to provide the complete set of images while preserving anonymity.

\section{Dataset Release}

We will release Fish-Vista on HuggingFace. We will also release a HuggingFace dataset card with a detailed description of the metadata, data instances, annotation, and license information. 

\subsection{Licensing Information}
The source images in our dataset come with various licenses, mostly within the Creative Commons family. We will provide license and citation information, including the source institution for each image, in our metadata CSV files, which will be made available in the HuggingFace repository. Additionally, we will attribute each image to the original FishAIR URL from which it was downloaded.

A small subset of our images (approximately 1k) from IDigBio are licensed under CC-BY-ND, which prohibits us from distributing processed versions of these images. Therefore, we will not publish these $\approx 1,000$ images. Instead, we will provide the URLs for downloading the original images and a processing script that can be applied to obtain the processed versions we use.

Our dataset will be licensed under CC-BY-NC 4.0. However, as mentioned earlier, individual images within our dataset have different licenses, which will be specified in our CSV files. We will provide the licenses of the original sources so that anyone using our dataset can adhere to the licensing requirements of the individual images.

\section{Further Details of Processing Pipeline}
\label{appendix:data-processing}
In this section, we provide further details of the  data processing pipeline that we use to obtain the images in Fish-Vista. 

\subsection{Examples of Raw Museum Images}

As mentioned in Section \ref{section:data-processing}, the raw images obtained from the FishAIR repository exhibit a range of noisy artifacts. We observe that images of museum specimens predominantly include rulers and tags (Figure \ref{fig:cropping-examples}, Raw Image). Some images contain radiographic images (Figure \ref{fig:discarded-by-detection}, first row), while others include hand-written notes with no fish images (Figure \ref{fig:discarded-by-detection}, second row). 

\subsection{Quality Metadata-based Filtering (Step 2)}

In this step, we leverage quality metadata provided in CSV files by Fish-AIR, containing manually annotated information about the image quality of museum fish specimens. The metadata include the following fields (among others):

\begin{itemize} \item \textit{allPartsVisible}: A boolean variable indicating whether all parts of the fish specimen are visible or not. \item \textit{partsMissing}: A boolean indicating whether any parts of the specimen are missing or not. \item \textit{specimenView}: A categorical variable specifying the view of the specimen (e.g., `top view', `bottom view', `side view', `complicated view'). \end{itemize}

At the time of this study, quality metadata were available for 29,075 GLIN images and 1,435 iDigBio images used in our dataset. No quality metadata were available for the Morphbank images.

\begin{figure}[ht]
    \centering
    \includegraphics[width=1\linewidth]{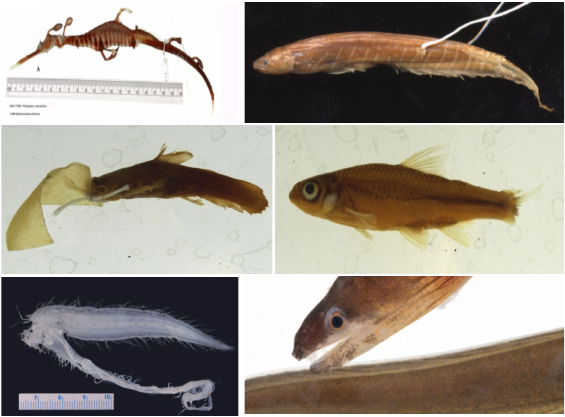}
    \caption{Examples of images filtered during the metadata filtering step of the processing pipeline.}
    \label{fig:metadata-filtered-images}
\end{figure}

We filtered out images based on the following criteria:

\begin{enumerate} \item Images where \textit{allPartsVisible} is \textit{False} (see Figure \ref{fig:metadata-filtered-images}, top row). \item Images where \textit{partsMissing} is \textit{True} (see Figure \ref{fig:metadata-filtered-images}, middle row). For example, the first image shows a specimen missing its head, and the second image is missing its tail. \item Images labeled with a \textit{specimenView} of  `complicated view'. Manual inspection revealed that these images do not adequately display the visual traits that we need to analyze (see Figure \ref{fig:metadata-filtered-images}, bottom row). \end{enumerate}

As a result of this filtering process, we discarded 4,467 images from GLIN and 301 images from iDigBio.

\subsection{Detecting and Cropping Fish Bounding Boxes (Step 4)}
We use Grounding DINO to detect and extract tight bounding boxes around fish specimens in the images. This step ensures that images without any fish specimens are excluded. Additionally, this process removes undesired visual elements, such as rulers and tags, which could otherwise introduce noise and detract machine learning models from focusing on the visual traits of the specimens. Additionally, museum images often feature multiple fishes within a single frame. To facilitate the study of visual traits, it is essential to ensure that each image contains only a single fish specimen. By detecting and cropping individual fishes, we achieve this objective, resulting in a dataset of individual fish images. 

\textbf{Grounding DINO implementation details:} Grounding DINO uses a textual prompt to detect bounding boxes in an image. For our use case, we find that using the prompt ``fish'' results in good detection of fish specimens from museum images. A \textit{box\_threshold} of 0.4 is set for initial detection, but only bounding boxes with a confidence score of 0.5 or higher are retained. We avoid setting a higher confidence threshold to minimize exclusion of valid fish images.

\begin{figure}[ht]
    \centering
    \includegraphics[width=1\linewidth]{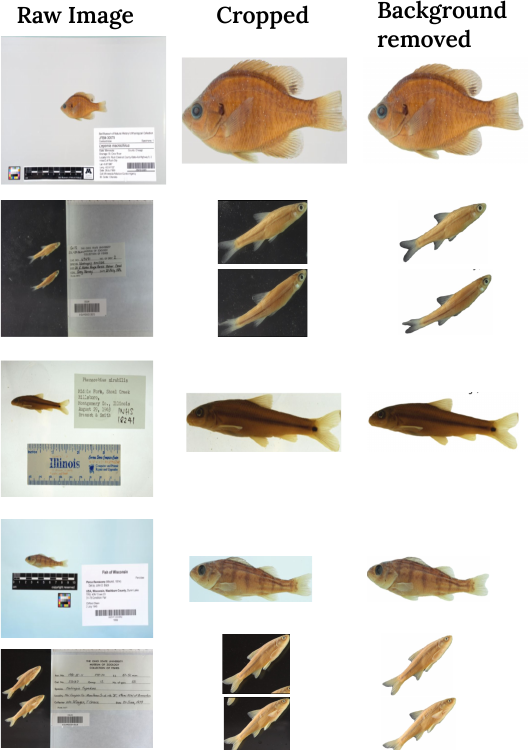}
    \caption{Examples of raw images from Fish-AIR (first column), crops generated by Grounding DINO (middle column) and background removed images by SAM (last column)}
    \label{fig:cropping-examples}
\end{figure}

\textbf{How good is Grounding DINO on detecting fish from museum images?} In order to validate the use of Grounding DINO, we manually inspected $\approx 500$ randomly chosen images and observed no inaccuracies in bounding box detection. We show some examples in Figure \ref{fig:cropping-examples}, second column. For quantitative evaluation, we obtained 311 GLIN museum fish images from \cite{pepper2021automatic}, which contains manually annotated bounding boxes of fishes. We obtained an mIOU of 90.1\%, which shows that our bounding boxes are tight.

\begin{figure}[ht]
    \centering
    \includegraphics[width=1\linewidth]{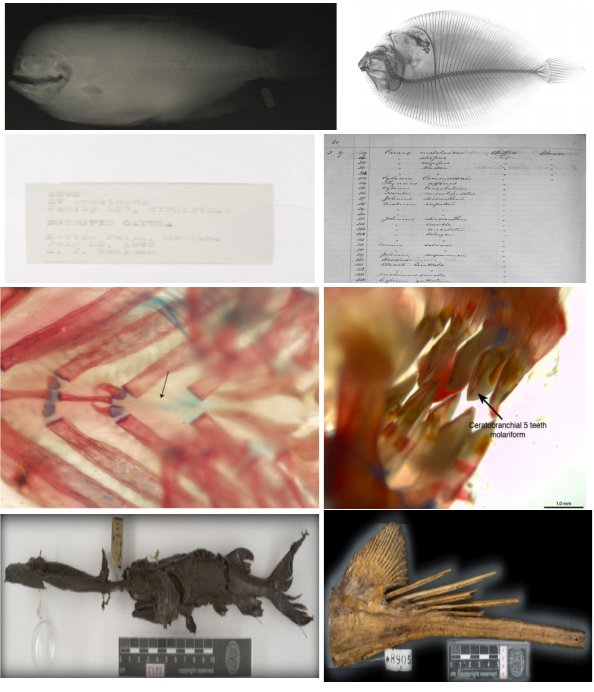}
    \caption{Examples of noisy images that are discarded by the fish detection step of Grounding DINO.}
    \label{fig:discarded-by-detection}
\end{figure}

Following the detection process, we discard 2,062 images where no fish specimens were detected. Figure \ref{fig:discarded-by-detection} shows a few examples of the discarded images. Since individual images may contain multiple fishes, our cropping approach results in the addition of 12,320 bounding boxes to the dataset, corresponding to individual fishes. To maintain a minimum quality standard, we further filter out bounding boxes with height and width smaller than 224 pixels, ensuring that very low-resolution images are excluded from our dataset. This step results in the removal of 422 bounding boxes from the dataset.

\subsection{Removing Background using SAM (Step 5)}
\textbf{Why do we remove backgrounds?} Museum collection images often feature artificial backgrounds, which can introduce unintended biases into trained models. For instance, if a particular species is consistently associated with a black background, while other species lack such backgrounds, the classifier may learn to distinguish backgrounds rather than focusing on the visual traits of the specimens. To mitigate this bias and create a controlled experimental environment, we remove backgrounds from all images.

We use the Segment Anything Model (SAM) to remove backgrounds from the fish images. Specifically, we use the bounding boxes from Step 4 (Grounding DINO) as prompts to SAM to detect foregrounds from images. 
We then replace the backgrounds with white color, while cropping into the segmented fish. We use default hyperparameters suggested in the SAM repository and the ViT-H SAM model. 

Background removal using SAM also serves as a filtering step in our pipeline, operating in two key ways. First, SAM may detect no foreground in an image. In such cases, we discard those images. Second, we apply an explicit filtering condition: if SAM detects more than one foreground component, the image is discarded. This strict condition addresses two issues. Multiple detected components may indicate the presence of undesired elements, such as rulers or tags, that are either on or in close proximity to the fish body, and therefore falls within the bounding box. Alternatively, it may indicate that the fish specimen is broken or disconnected, suggesting deformation that we aim to exclude from the dataset.

This filtering step removes approximately 12,000 images, resulting in a final dataset of 100,300 images spanning 10,681 species. 

\textbf{How good is SAM on segmenting whole fishes and the background?} In order to validate the use of SAM, we manually inspected $\approx 500$ randomly chosen images and observed no inaccuracies in the background removal. We show some examples in Figure \ref{fig:cropping-examples}, third column. For quantitative evaluation, we utilized manual segmentation annotations for 492 whole fish images sourced from GLIN, provided by \cite{pepper2021automatic}. Using these annotations as ground truth, SAM achieved an mIoU of 90.8\%, demonstrating its capability to accurately segment fish specimens. These results confirm the suitability of SAM for background removal in our dataset.

\section{Manual Filtering of Species}
\label{appendix:manual-species-filtering}
After completing the data processing steps detailed in Section \ref{section:data-processing} and Appendix \ref{appendix:data-processing}, we obtain complete fish images free from noisy artifacts and with uniform backgrounds. However, further filtering is required to remove images that may not adequately exhibit visual traits. This issue arises when fish specimens are photographed in views where traits are obscured or when specimens are deformed due to prolonged preservation in museum conditions.

\begin{figure}
    \centering
    \includegraphics[width=1\linewidth]{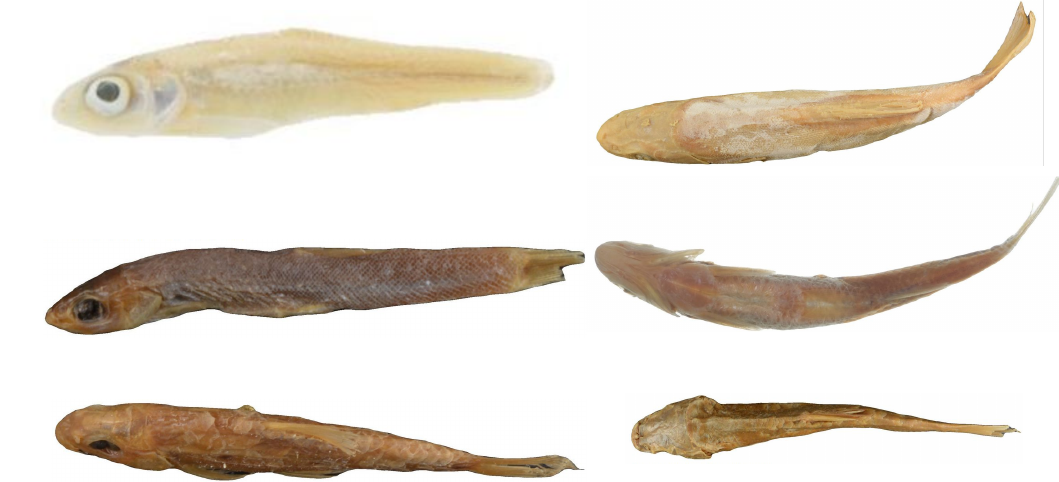}
    \caption{Examples of images that do not demonstrate visual traits. We consider such images to be of bad quality during our manual inspection and filtering.}
    \label{fig:bad-images}
\end{figure}

To address this, we perform a manual inspection of the remaining images. Our inspection follows a rule-of-thumb: an image is deemed low quality if any of the key traits, such as the eye, tail, or head, are not visible, or if fewer than two fins are visible. Examples of filtered images from this process are shown in Figure \ref{fig:bad-images}. These examples clearly demonstrate the absence of visual traits, justifying their removal to maintain dataset quality.

Given the labor-intensive nature of manual inspection, filtering every image in the dataset is infeasible. Instead, we randomly sample 15\% of images per species for manual inspection. If more than half of the sampled images for a species meet the criteria for being filtered out, we infer that most images of that species are of low quality and discard the entire species from the dataset. This approach primarily impacts species with fewer images per species, which are more prone to containing a significant proportion of low-quality images. In total, we discard 420 species, comprising a total of 4,886 images, during this manual filtering step.
Also, this filtering is applied only to obtain the classification and identification datasets but not the segmentation dataset, as the segmentation dataset is entirely manually annotated. 

While some low-quality images may still remain in the dataset, we expect that the majority of images in our dataset are of good quality that can be used for training machine learning models. To guarantee clean evaluation, every image in the test sets are manually inspected. Noisy images are discarded during this process, as detailed in Appendix \ref{appendix:data-split}, ensuring that the test sets remain free of noisy images and do not negatively impact the evaluation of model performance.

\section{Manual Annotation for Segmentation}

\begin{figure}[ht]
    \centering
    \includegraphics[width=1\linewidth]{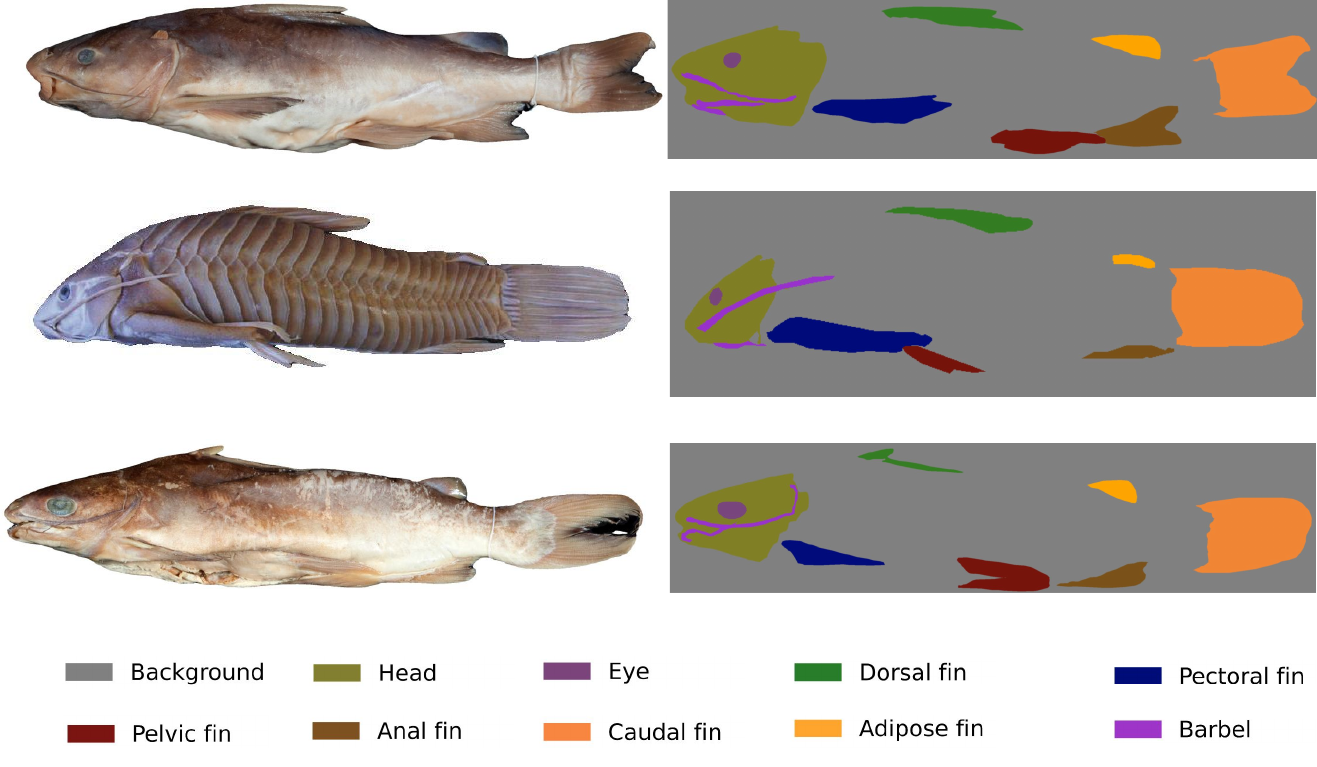}
    \caption{Examples of annotated visual trait segmentations for the nine different traits from the Segmentation dataset.}
    \label{fig:segmentaion-annotation-examples}
\end{figure}

12 annotators used the Computer Vision Annotation Tool (CVAT) \cite{cvat2023} to annotate nine traits in a subset of processed images: eye, head, barbel, dorsal fin, adipose fin, pectoral fin, pelvic fin, anal fin, and caudal fin. We provide additional examples of the annotations in Figure \ref{fig:segmentaion-annotation-examples}. These traits were chosen due to their well-defined physical boundaries which can be accurately segmented using CVAT. 

We prioritized images containing specimens oriented in lateral view over specimens in top or bottom-view for consistency and to maximize the visibility of traits. Images of damaged or degraded specimens were excluded (similar to those shown in Figure \ref{fig:bad-images}), as were images with poor resolution. We also omitted images of specimens that are difficult to photograph in standard lateral view, such as elongated species or those prone to curling when preserved (e.g., eels).

\section{Additional Dataset Details}
\label{appendix:data-split}
\subsection{Classification Dataset}

To create the classification dataset, we further filter the dataset by retaining only species with at least 4 images per species for the classification dataset. This ensures that the classification dataset includes a minimum of 2 images for training, 1 for testing, and 1 for validation for every species. Following this step, and the manual test-set filtering step described below, the final classification dataset consists of 56,360 images spanning 1,758 species.

In order to create the train, test, and validation split, we perform a stratified split of $75\%$, $15\%$ and $10\%$ respectively. We set a minimum threshold of 1 image in the test set and 1 image in the validation set for cases where the splitting would result in no images being set out for the test and validation splits. This results in a training set of 39,800 images, a validation set of 6,779 images and an \textit{initial test set} of 10,830 images.

\textbf{Manual filtering of the initial test set: }In order to ensure that the models are tested on a perfectly clean dataset, we manually inspect every image in the initial test set to obtain our final test set for classification. We follow the same manual inspection guidelines as discussed in Appendix \ref{appendix:manual-species-filtering}. We discard 1,049 images that do not have traits visible, either owing to deformity, or because of the view of the image. This is less than 10\% of the {initial test set} images, which demonstrates the effectiveness of our prior data processing steps. We obtain a \textit{final test set} of 9,781 images.

We provide statistics across the four categorizations of species: majority, neutral, minority and ultra-rare, along with an overview of the long-tailed distribution in Figure \ref{fig:key_stat} - Species Classification. 

\subsection{Identification Dataset}
The trait identification dataset is designed to achieve three key objectives: detecting the presence or absence of four traits—adipose fin, pelvic fin, barbel, and multiple dorsal fins; evaluating model performance on out-of-distribution (OOD) test sets; and assessing whether traits can be localized in images using coarse-grained weak labels by predicting their presence/absence. The initial dataset, consisting of 52,982 images from 682 species, is divided into four splits: training, validation, in-species test, and leave-out-species test sets. We reserve 51 species (1,935 images) for the leave-out test set, ensuring sufficient variation in the presence and absence of all four traits for robust OOD evaluation. The remaining species are split into training (75\%), in-species test (15\%), and validation (10\%) sets. The training set comprises 38,038 images from 628 species, the validation set includes 5,238 images from 451 species, and the in-species test set contains 7,771 images from 450 species, all of which overlap with the species in the training set to provide an in-distribution evaluation set. To ensure high-quality evaluation, we manually inspect the test sets to remove noisy or low-quality samples, following the same process used for the classification dataset. 

We construct a manual-annotation dataset comprising presence/absence annotations for four traits across 1,281 images spanning 1,075 species. This dataset is derived from a subset of the manually annotated segmentation dataset, carefully curated to ensure no overlap with the other four identification datasets and no species overlap with the training set. For this dataset, the presence/absence of traits such as the adipose fin, pelvic fin, and barbel is straightforward to infer from segmentation annotations, as the presence of corresponding pixel labels directly indicates the traits' presence. However, this approach cannot be applied to images with multiple dorsal fins, as all dorsal fins share the same pixel label in the segmentation annotations. To address this, we used help from expert biologists in our team to manually inspect these images and annotate the presence/absence of multiple dorsal fins. 

Incorporating the \textit{manual-annotation} dataset brings the total identification dataset to 54,263 images spanning 1,754 species. The key statistics of the identification dataset are illustrated in Figure \ref{fig:key_stat} - Trait Identification. In the figure, the presence percentage for each trait represents the proportion of images in which the trait is present, highlighting the highly imbalanced distribution of each trait in our dataset. 

\subsection{Segmentation Dataset}

\begin{table}[ht]
\resizebox{0.48\textwidth}{!}{%
\begin{tabular}{lccccccccc}
            & Head & Eye  & Dorsal & Pectoral & Pelvic & Anal  & Caudal & Adipose & Barbel \\ 
            \toprule
Presence \% & 99.5 & 97.7 & 97.58  & 98.46    & 94.02  & 98.05 & 98.65  & 9.40    & 8.12   \\
Avg Area \% & 6.88 & 0.71 & 4.83   & 2.49     & 1.03   & 2.60  & 5.00   & 0.42    & 0.47 \\
\bottomrule
\end{tabular}%
}
\caption{\small{The proportion of images where each of the nine traits are present, and the average area they occupy per image}}
\label{tab:segmentation-trait-stats}
\end{table}

For the segmentation dataset, we create a split of \textit{70\%-25\%-5\%} split for train-test-validation sets respectively, stratfied by the unique combination of the nine traits. We obtain a training set of 4,102 images, test set of 1,446 images and validation set of 290 images. Key statistics are shown in Figure \ref{fig:key_stat}. In this figure, the `Presence \%' indicates the proportion of images for which a trait is present, and the `Avg Area \%' indicates the average proportion of pixels that a trait covers. The complete table of Presence \% and Avg Area \% for all 9 traits are provided in Table \ref{tab:segmentation-trait-stats}. We can see the high imbalance associated with the trait presences, as well as the very small area covered by some of the traits, particularly the eye, barbel and the adipose fin. 

\section{Experiment Details}

\subsection{Classification Experiments}
\label{sec:classification_experiment_details}
\textbf{Hyperparameters: } For all the CNN-based backbone models reported in Table \ref{tab:classification}, we use hyperparameters following suggestions of training routines for imbalanced image datasets provided in \cite{kim2020m2m} and \cite{shwartz2024simplifying}. We use the SGD optimizer with a base learning rate of 0.1, with a linear warmup of the learning rate for 5 epochs.  We also employ cosine annealing decay for the learning rate. We use a weight decay parameter of 2e-4.  We train all CNN-based models for 100 epochs, since we observe that these models converge well within this limit. We employ early stopping with a patience of 10 epochs that goes into effect after training for the first 50 epochs. 

For all the transformer-based backbone models, we use hyper-parameter suggestions from \cite{steiner2021train}. We use Adam optimizer with base learning rate of 3e-4, and a linear warm-up of the learning rate for the first 50 epochs. We set the weight decay parameter to be 0.1. We train the transformer-based models for 150 epochs, since we observe that these models take longer to converge than the CNN-based models. For all classification experiments, we use a batch size of 128. We also employ cosine annealing decay for the learning rate. All of our models are pretrained on the ImageNet-1k \cite{imagenet} dataset, unless explicitly tagged with \textit{22k}, in which case we start with ImageNet-22k \cite{imagenet} pretrained weights. We employ early stopping with a patience of 10 epochs that goes into effect after training for the first 100 epochs.

\textbf{Image augmentations: }  For species classification and trait identification, it is essential to maintain the aspect ratio of the various parts of the input images. Therefore, we pad all images along the shorter edge to make both sides of the image the same length (square padding). We then resize the image to the required resolution while maintaining the aspect-ratio according to the model that we use. For the classification baselines, we use a resolution of 224 $\times$ 224 with an exception of Inception-v3, Mobile\_ViT and EfficientNet-v2, which use resolutions of 299 $\times$ 299, 256 $\times$ 256, and 480 $\times$ 480 respectively. 
We calculate the mean and standard deviation of our training set and normalize every input image accordingly. 
During training, we randomly augment the images with the following operations: rotations between 0 and 180 degrees, adjusting the sharpness, changing the contrast, and performing horizontal and vertical flips. 

\textbf{Loss function: } For all trained methods, we use the standard cross entropy (CE) loss with the objective of predicting the correct class through empirical risk minimization.

\subsubsection{Details of Zero-Shot Classification Experiments}
For CLIP and Bio-CLIP Zero-Shot (ZS), we use textual prompt ensembling using the same set of 80 prompts provided by OpenAI in the original CLIP paper \cite{radford2021learning}. 

\subsubsection{Details of Fine-grained Categorization Methods}

We experiment using two recent FGVC methods --  INTR and TransFG (see Table \ref{tab:classification}). We use their default settings of hyper-parameters and follow the implementations as provided in the original repositories. We refer readers to the original papers for the model architectures. 
While TransFG demonstrates reasonable performance on Fish-Vista, INTR exhibits the poorest performance among all models on the minority and ultra-rare species. This observation indicates that fine-grained visual categorization methods may struggle to handle the imbalanced long-tailed distribution of our dataset effectively.

\subsubsection{Details of Imbalanced Methods}

\textbf{Class-balanced re-weighting (CB-RW): } CB-RW \cite{cui2019class} is a re-weighting strategy that assigns weights to each class based on the inverse of the effective number of samples in the class. The effective number is defined as a function of the number of samples in class \( k \), denoted as \( N_k \), and a hyperparameter \( \beta \). The weight for class \( k \) is given by:

\begin{equation}
w_k = \frac{1 - \beta^{N_k}}{1 - \beta}
\end{equation}

For our experiments, we set \( \beta = 0.9999 \). This weighting scheme ensures that underrepresented classes receive higher weights, addressing the class imbalance problem.

\textbf{Focal Loss: } Focal loss \cite{lin2017focal} down-weights the loss for well-classified examples, thus reducing their impact during training. By applying a modulation factor to the standard cross-entropy loss, focal loss ensures the model concentrates on difficult, underrepresented classes. In our implementation, we use $\gamma=2$ for the loss modulation factor. 


\begin{table}[ht]
\setlength\tabcolsep{0.08cm} 
\renewcommand{\arraystretch}{0.9}
\resizebox{0.48\textwidth}{!}
{%
\begin{tabular}{lccccc}
\toprule
        
        \fontsize{9.2pt}{10}\selectfont\textbf{Model} & \fontsize{9pt}{10}\selectfont\textbf{F1} & \begin{tabular}[c]{@{}c@{}}{\fontsize{9pt}{10}\selectfont\textbf{Major}} \\
        {\fontsize{9pt}{10}\selectfont\textbf{Acc.}}\end{tabular} & \begin{tabular}[c]{@{}c@{}}{\fontsize{9pt}{10}\selectfont\textbf{Neutral}} \\ {\fontsize{9pt}{10}\selectfont\textbf{Acc.}}\end{tabular} & \begin{tabular}[c]{@{}c@{}}{\fontsize{9pt}{10}\selectfont\textbf{Minor}} \\ {\fontsize{9pt}{10}\selectfont\textbf{Acc.}}\end{tabular} & \begin{tabular}[c]{@{}c@{}}{\fontsize{9pt}{10}\selectfont\textbf{Ultra-R}} \\
        {\fontsize{9pt}{10}\selectfont\textbf{Acc.}} \end{tabular}\\ \midrule
 \fontsize{8.5pt}{10}\textbf{VGG-19}\cite{simonyan2014very}             & 49.7                     & \colorbox{blue!15}{93.5} & 83.0                     & 74.2                     & 45.9                     \\
 \fontsize{8.5pt}{10}\textbf{Resnet-34} \cite{resnet}           & \colorbox{red!15}{35.6}  & 89.9                     & \colorbox{red!35}{68.4}  & \colorbox{red!35}{60.9}  & \colorbox{red!15}{30.7}  \\
 \fontsize{8.5pt}{10}\textbf{Inception-v3} \cite{inceptionv3}       & 40.2                     & 90.0                     & 77.7                     & 67.7                     & 34.5                     \\
 \fontsize{8.5pt}{10}\textbf{ResNext-50} \cite{xie2017aggregated}    & 44.4                     & 91.4                     & 78.3                     & 69.8                     & 39.1                     \\
 \fontsize{8.5pt}{10}\textbf{MobileNet-v3} \cite{howard2019searching} & 40.1                     & \colorbox{red!35}{86.0}  & \colorbox{red!15}{74.4}  & 65.5                     & 34.0                     \\
 \fontsize{8.5pt}{10}\textbf{RegNet-y} \cite{radosavovic2020designing}          & 43.7                     & 89.8                     & 77.4                     & 68.5                     & 38.5                     \\
 \fontsize{8.5pt}{10}\textbf{EfficientNet-v2} \cite{tan2021efficientnetv2}  & \colorbox{red!35}{34.3}  & 89.0                     & 75.0                     & \colorbox{red!15}{62.3}  & \colorbox{red!35}{28.5}  \\
 \fontsize{8.5pt}{10}\textbf{ConvNext-B} \cite{convnext}  & 49.5  & 89.6                     & 81.8                     & 73.1  & 44.9  \\
 \midrule
 \fontsize{8.5pt}{10}\textbf{ViT-B-16} \cite{ViT}          & 48.3                     & 88.7                     & 82.3                     & 73.3                     & 43.4                     \\
 \fontsize{8.5pt}{10}\textbf{ViT-B-32} \cite{ViT}          & 45.2                     & \colorbox{red!15}{86.9}  & 75.8                     & 66.6                     & 41.8                     \\
 \fontsize{8.5pt}{10}\textbf{DEiT-distilled-s} \cite{touvron2021training}  & 46.2                     & 91.7                     & 76.8                     & 72.3                     & 40.8                     \\
 \fontsize{8.5pt}{10}\textbf{Swin-B-22k} \cite{swin}         & \colorbox{blue!15}{55.1} & 92.6                     & \colorbox{blue!15}{86.2} & \colorbox{blue!15}{79.6} & \colorbox{blue!15}{50.4} \\
 \fontsize{8.5pt}{10}\textbf{CVT-13} \cite{cvt}            & 49.3                     & 92.0                     & 83.3                     & 73.5                     & 44.7                     \\
 \fontsize{8.5pt}{10}\textbf{MobileViT-xs} \cite{mehta2021mobilevit}     & 49.0                     & 92.2                     & 85.9                     & 74.1                     & 43.7                     \\
 \fontsize{8.5pt}{10}\textbf{MobileViT-v2} \cite{mehta2021mobilevit}     & 42.7                     & 91.4                     & 80.8                     & 66.8                     & 37.6                     \\
 \fontsize{8.5pt}{10}\textbf{MaxViT-t} \cite{maxvit}          & \colorbox{blue!35}{57.8} & \colorbox{blue!35}{94.4} & \colorbox{blue!35}{86.7} & \colorbox{blue!35}{81.4} & \colorbox{blue!35}{53.9} \\
 \fontsize{8.5pt}{10}\textbf{PVT-v2} \cite{wang2022pvt}            & 51.0                     & 92.0                     & 83.4                     & 75.7                     & 45.8  \\
 \bottomrule
\end{tabular}%
}
\caption{\small{Comparison of the classification performance  (in \%) of different mainstream CNN-based and transformer-based vision backbones. Results are color-coded as \colorbox{blue!35}{Best}, \colorbox{blue!15}{Second best}, \colorbox{red!35}{Worst}, \colorbox{red!15}{Second worst}.}}.
\label{tab:classification-appendix}
\end{table}


\subsection{Identification Experiments}

\textbf{Hyperparameters: } For all the backbone models used in trait identification, we use the Adam optimizer with weight decay of 0.1. We train every model for 150 epochs. For CNN-based models, we use a maximum learning rate of 1e-4 with a linear warm-up for 5 epochs. For transformer-based models, we use a maximum learning rate of 3e-4 with a linear warm-up of 50 epochs. We use cosine annealing learning rate decay. We train with a batch size of 128. 

For the Query2Label \cite{liu2021query2label} models, we use the default set of hyperparameters used in the original paper. We use the Adam optimizer with weight decay coefficient of 1e-2. We use a learning rate of 1e-4 with cosine annealing. In the Query2Label transformer, we use 1 encoder layer and 2 decoder layers. We vary the number of heads between 1 and 4. We use Resnet34 and SWIN-base backbones, pretrained on the ImageNet-22k dataset

\textbf{Image augmentations: }  We use the same augmentations for the identification experiments that we use for classification, described in Section \ref{sec:classification_experiment_details}.

\textbf{Loss function: }Binary cross entropy loss is used to train the models, since we have a multi-label classification objective.  In order to account for the imbalance demonstrated by each trait (Figure \ref{fig:key_stat}), we use the weighted binary cross entropy loss for all models except Query2Label. For each trait, the loss for minority labels is scaled by a factor $\Gamma_{scale}$, where $\Gamma_{scale} = \frac{N_{major}}{N_{minor}}$ and $N_{major}, N_{minor}$ are the number of majority labels and minority labels for each trait, respectively. Query2Label uses the assymmetric loss as part of their implementation, and we use the default implementation presented in the original paper. 

\subsubsection{Attention Maps from Query2Label}
We visualize the attention maps shown in Figure \ref{fig:attention-q2l-swin} following the method described in the original Query2Label paper. 

For multi-head models, we take the mean of the multiple attention maps. The attention maps are interpolated to the original image size. This allows us to compute the mIoU with the ground-truth segmentation maps on the \textit{manual-annotation} test set, as shown in Table \ref{tab:attention-miou}. Since the model is trained on squared images, we ensure that the attention maps are interpolated according to the resized input image.

\subsection{Segmentation Experiments}

\textbf{Hyperparameters: }
For the semantic segmentation methods listed in the first section of Table \ref{tab:segmentation} (PSPNet to Semantic FPN), we use the implementation provided in the Segmentation Models Pytorch (SMP) library. For our experiments, we used the Adam optimizer with a learning rate of 2e-4. The learning rate was scheduled using a cosine annealing learning rate scheduler, with a minimum learning rate of 1e-5. The models were trained with a batch size of 32 for up to 100 epochs. Early stopping was employed with a patience of 10 epochs. 

For the instance segmentation methods listed in the second section of Table \ref{tab:segmentation} (Mask2Former and YOLOv8), we used a learning rate of 2.5e-4 and a batch size of 4. 

\textbf{Augmentations: }We used the \textit{albumentations} library in pytorch for training data augmentations. The augmentation pipeline includes horizontal flipping with a probability of 0.5 and shift-scale-rotate transformations that allow scaling up to 50\%, rotating within a limit of 0 degrees, and shifting up to 10\%, applied with a probability of 1. The images were resized to a maximum size of 320 pixels while maintaining the aspect ratio, and padding was added as needed to ensure a size of 320 $\times$ 320 pixels. Padding used a constant border mode with a white background. Gaussian noise was added to images with a probability of 0.2, and perspective transformations were applied with a probability of 0.5. To enhance brightness and contrast variations, one of the following augmentations was randomly applied with a probability of 0.9: CLAHE (Contrast Limited Adaptive Histogram Equalization), random brightness and contrast adjustment, or gamma adjustment. The pipeline also included blurring effects, where one of the following was applied with a probability of 0.9: sharpening, Gaussian blur, or motion blur, each with a blur limit of 3. Furthermore, to introduce color variations, one of the following was randomly applied with a probability of 0.9: hue-saturation adjustment or additional brightness and contrast adjustment. This comprehensive augmentation strategy was adapted from default recommendations in the SMP library.

\vspace{-20pt}
\subsubsection{Molmo-SAM Implementation Details}

For the zero-shot segmentation method combining Molmo and SAM, we provide Molmo with images in the segmentation test set and prompt it using the text: ``Point me to the $<trait>$ of the fish." The placeholder $<trait>$ is replaced with one of the nine trait names listed in Figure \ref{fig:segmentaion-annotation-examples}. For the caudal fin, $<trait>$ is replaced with ``caudal fin or tail" to account for the non-scientific terminology, as the caudal fin is commonly referred to as the tail.

Molmo outputs numeric points corresponding to the traits, if detected. Using these points, we prompt SAM-v2 to generate nine binary segmentation masks for each image, where each mask corresponds to one of the nine traits. These binary masks are then merged into a single segmentation map labeled with the different traits. In cases where traits overlap, we resolve the conflict using a predefined priority order (low to high): {Head, Eye, Dorsal Fin, Pectoral Fin, Pelvic Fin, Anal Fin, Caudal Fin, Adipose Fin, Barbel}. Traits with higher priority are assigned overlapping pixels. This priority order is determined based on the physical arrangement of traits and their segmentation difficulty. For instance, Eye is prioritized over Head since the eye is always within the head, while Adipose Fin and Barbel are given the highest priority as they are the most challenging traits to segment.

In our implementation, we use the `allenai/Molmo-7B-D-0924' variant of Molmo. We enable greedy decoding (we set temperature to 0), to prevent varying outputs. We use the `sam2.1\_hiera\_large' variant of SAM-v2, and use default configuration provided in the SAM-v2 repository. 

Exploring alternative methods to enhance the performance of the Molmo+SAM pipeline is an interesting direction for future work but is beyond the scope of this paper.

\section{Additional Results}

\subsection{Classification}
In addition to the classification results provided in the main paper in Table \ref{tab:classification}, we provide a comprehensive evaluation of mainstream vision backbones in Table \ref{tab:classification-appendix}.

\subsection{Identification}
A comprehensive benchmarking was conducted for trait identification, evaluating each model on the three evaluation sets: the \textit{in-species test set} (Table \ref{tab:in-species-test-set}), the \textit{leave-out-species test set} (Table \ref{tab:leave-species-out-test-set}), and the \textit{manual-annotation test set} (Table \ref{tab:manually-annotated-test-set}). In terms of the metrics, we report the mean average precision (mAP), the average precision for each of the four traits, the macro-averaged F1 score at a 0.5 threshold, and the macro-averaged F1 score at the optimal threshold. For each model, the optimal threshold is determined from the precision-recall curve of the validation set.

The results show high performance on the \textit{in-species test set}, with progressively lower performance on the \textit{leave-out-species test set} and the \textit{manual-annotation test set}. Computing the optimal threshold for F1 score generally improves performance over the default 0.5 threshold, which is expected given the imbalanced nature of our traits. Notably, all variants of the Query2Label model -- Resnet34 (R34) and SWIN backbones, each traned with either a single head (SH) or four heads (Multiple Head, MH) -- consistently outperforms other models. However, even Query2Label faces significant challenges in achieving decent performance on the challenging \textit{manual-annotation} set. This underscores the difficulty state-of-the-art models encounter in robustly generalizing to predict the fine-grained visual traits.

\begin{figure*}[h]
    \centering
    \includegraphics[width=1\linewidth]{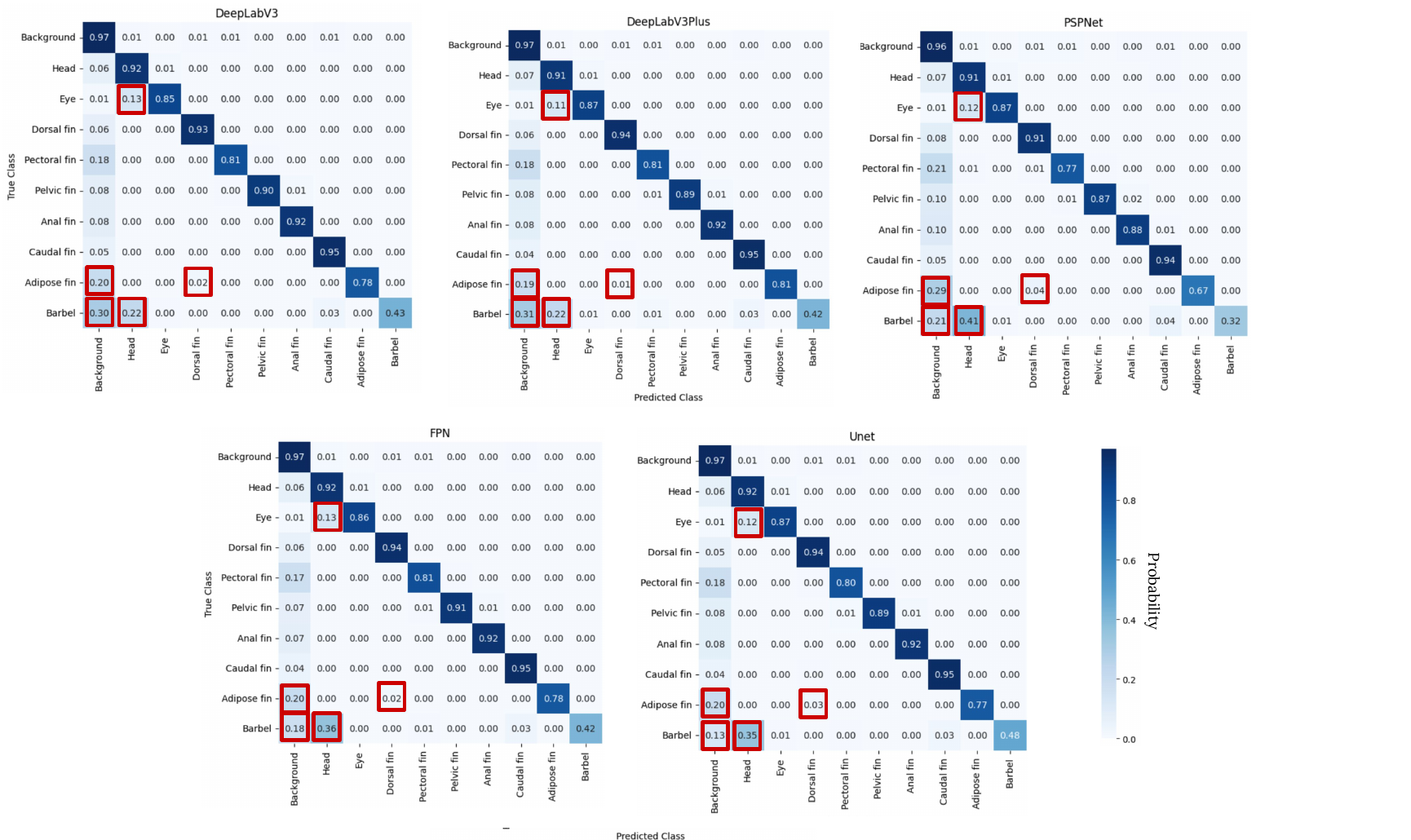}
    \caption{Confusion matrix for the five semantic segmentation models, with cells of interest highlighted in red frames.}
    \label{fig:confusion_analysis}
\end{figure*}

\subsection{Segmentation}
We present the confusion matrices for the five semantic segmentation models used in our experiments in Figure \ref{fig:confusion_analysis}, with cells of interest highlighted in red. A consistent pattern emerges across all models: the adipose fin, when misclassified, is often segmented as background, likely due to its rarity in the dataset. In other instances, it is misclassified as the dorsal fin, which can be attributed to their close spatial proximity, and their subtle difference in appearance, as shown in Figure \ref{fig:segmentaion-annotation-examples}.

Similarly, the barbel is frequently misclassified as background or as the head. This behavior can be explained by the barbel's rarity and its typical appearance over the head region. On the other hand, the eye, which is consistently present in our dataset and located on the head, is rarely misclassified as background. However, it is often segmented as part of the head due to their close association.

This analysis highlights key challenges in segmentation: small and rare traits are more likely to be segmented as background, traits that are spatially adjacent, or overlaid on other traits, or appear similar to other traits are prone to being misclassified as the other trait. These findings highlight the need for segmentation methods to handle rare, small, spatially nearby and fine-grained traits effectively. 




\begin{table*}[h]
\centering
\renewcommand{\arraystretch}{0.9}
\resizebox{\textwidth}{!}{
\begin{tabular}{lccccccccccccc}
\toprule

\multirow{2}{*}{{Model}}     & \multicolumn{5}{c}{{Average Precision}}                                                                                 & \multicolumn{4}{c}{{F1@0.5}}                                                                    & \multicolumn{4}{c}{{F1@optimal threshold}}                                                      \\ 
\cmidrule(lr){2-6} \cmidrule(lr){7-10} \cmidrule(lr){11-14}
                             & mAP    & Adip  & Pelv  & Barb  & Dors  & Adip  & Pelv  & Barb  & Dors  & Adip     & Pelv     & Barb     & Dors    \\ 
 \midrule
VGG-19             & 87.06                     & 95.09                     & 61.85                     & 96.21                     & 95.07                     & 92.61                     & 68.28                     & 94.62                     & 93.17                     & 93.69                     & 80.85                     & 94.7                      & 93.28                     \\
ResNet-18          & 87.85                     & 96.56                     & 65.96                     & 95.16                     & 93.73                     & 91.26                     & 78.95                     & 93.7                      & 91.16                     & 94.47                     & 79.01                     & 94.07                     & 92.18                     \\
ResNet-34          & 91.77                     & 95.71                     & 80.73                     & 95.53                     & 95.1                      & 94.42                     & 72.4                      & 94.64                     & 92.96                     & 94.41                     & 88.84                     & 94.72                     & 92.78                     \\
Inception-v3       & \colorbox{red!15}{77.0}   & 93.08                     & 52.4                      & \colorbox{red!35}{86.03}  & \colorbox{red!35}{76.47}  & \colorbox{red!35}{87.45}  & 77.67                     & \colorbox{red!15}{88.93}  & \colorbox{red!35}{83.53}  & 93.16                     & 78.48                     & \colorbox{red!35}{88.69}  & \colorbox{red!35}{84.95}  \\
ResNext-50         & 91.44                     & 98.53                     & 74.64                     & 97.03                     & 95.54                     & 96.07                     & 86.56                     & 96.32                     & 94.09                     & 96.9                      & 85.23                     & 96.45                     & 94.99                     \\
MobileNet-v3       & 90.49                     & 96.34                     & 73.79                     & 96.33                     & 95.51                     & 94.38                     & 83.73                     & 95.19                     & 93.69                     & 94.58                     & 84.26                     & 95.14                     & 94.73                     \\
RegNet-Y           & 89.51                     & 95.54                     & 73.06                     & 95.45                     & 94.0                      & 94.62                     & 82.28                     & 94.86                     & 94.12                     & 94.69                     & 81.6                      & 94.77                     & 94.04                     \\
EfficientNet-v2    & 95.96                     & 99.7                      & 86.41                     & 98.43                     & 99.3                      & 98.13                     & 90.69                     & 97.29                     & 98.04                     & 98.48                     & 92.22                     & 97.23                     & 98.28                     \\
ConvNext-Base      & 97.46                     & 99.54                     & 92.33                     & 98.67                     & 99.32                     & 98.62                     & 92.31                     & \colorbox{blue!15}{98.28} & 98.46                     & 98.62                     & 95.93                     & \colorbox{blue!15}{98.19} & 98.45                     \\
\midrule
ViT-B-16           & 86.95                     & 92.93                     & 67.7                      & 93.29                     & 93.89                     & 91.51                     & 66.75                     & 92.41                     & 92.57                     & \colorbox{red!15}{91.71}  & 81.23                     & 92.14                     & 92.68                     \\
ViT-B-32           & 81.44                     & \colorbox{red!35}{89.79}  & 59.81                     & \colorbox{red!15}{89.18}  & \colorbox{red!15}{86.97}  & \colorbox{red!15}{88.13}  & \colorbox{red!35}{64.63}  & \colorbox{red!35}{88.48}  & \colorbox{red!15}{86.57}  & \colorbox{red!35}{90.08}  & 79.07                     & \colorbox{red!15}{89.46}  & \colorbox{red!15}{87.91}  \\
DEiT-distilled     & 93.74                     & 97.61                     & 82.85                     & 97.02                     & 97.48                     & 95.5                      & 76.11                     & 94.93                     & 96.12                     & 95.72                     & 86.88                     & 95.76                     & 96.47                     \\
SWIN-B             & 92.02                     & 96.64                     & 76.98                     & 97.35                     & 97.11                     & 94.99                     & 82.12                     & 95.69                     & 95.46                     & 95.28                     & 86.78                     & 95.76                     & 95.76                     \\
SWIN-B-22k         & 95.21                     & 98.05                     & 86.94                     & 97.55                     & 98.28                     & 95.95                     & 86.37                     & 96.07                     & 96.47                     & 96.12                     & 91.29                     & 96.13                     & 96.62                     \\
CVT-13             & 83.61                     & 93.22                     & \colorbox{red!15}{49.56}  & 96.21                     & 95.43                     & 92.92                     & 71.72                     & 94.47                     & 94.08                     & 93.37                     & \colorbox{red!15}{78.1}   & 94.47                     & 94.61                     \\
Mobile-ViT-xs      & \colorbox{red!35}{75.12}  & \colorbox{red!15}{90.57}  & \colorbox{red!35}{25.66}  & 93.78                     & 90.5                      & 91.94                     & \colorbox{red!15}{65.17}  & 93.84                     & 91.21                     & 92.7                      & \colorbox{red!35}{66.94}  & 94.25                     & 90.96                     \\
Mobile-ViT-v2      & 86.3                      & 95.21                     & 62.3                      & 95.29                     & 92.4                      & 92.81                     & 70.38                     & 94.72                     & 91.66                     & 93.28                     & 78.25                     & 94.38                     & 92.32                     \\
MaxViT-t           & 95.49                     & 98.33                     & 86.44                     & 98.88                     & 98.3                      & 96.87                     & 87.71                     & 97.91                     & 97.77                     & 97.06                     & 91.99                     & 97.74                     & 97.68                     \\
PVT-v2             & 96.48                     & 98.84                     & 90.85                     & 97.75                     & 98.49                     & 97.61                     & 84.15                     & 96.94                     & 97.58                     & 97.62                     & 91.63                     & 97.15                     & 97.92                     \\
\midrule
Q2L (R34-22k SH) \cite{liu2021query2label}  & 97.78                     & 99.47                     & 93.91                     & 98.47                     & 99.26                     & 97.65                     & \colorbox{blue!15}{97.36} & 97.35                     & 98.27                     & 97.9                      & 95.1                      & 97.73                     & 98.45                     \\
Q2L (R34-22k MH)   & 97.22                     & 99.4                      & 91.92                     & 98.38                     & 99.17                     & 98.05                     & 96.65                     & 97.43                     & 98.39                     & 98.36                     & \colorbox{blue!15}{96.03} & 97.61                     & 98.62                     \\
Q2L (SwinB-22k SH) & \colorbox{blue!35}{98.32} & \colorbox{blue!15}{99.81} & \colorbox{blue!35}{94.94} & \colorbox{blue!35}{99.0}  & \colorbox{blue!15}{99.53} & \colorbox{blue!35}{99.09} & \colorbox{blue!15}{97.36} & 97.82                     & \colorbox{blue!35}{99.19} & \colorbox{blue!35}{98.93} & 95.66                     & 98.03                     & \colorbox{blue!15}{99.06} \\
Q2L (SwinB-22k MH) & \colorbox{blue!35}{98.32} & \colorbox{blue!35}{99.82} & \colorbox{blue!15}{94.91} & \colorbox{blue!15}{98.94} & \colorbox{blue!35}{99.61} & \colorbox{blue!15}{98.73} & \colorbox{blue!35}{98.04} & \colorbox{blue!35}{98.38} & \colorbox{blue!15}{99.02} & \colorbox{blue!35}{98.93} & \colorbox{blue!35}{96.82} & \colorbox{blue!35}{98.35} & \colorbox{blue!35}{99.23} \\
\bottomrule 
\end{tabular}%
}
\caption{Trait identification results on mainstream visual models using the \textbf{in-species test set}. Results are color-coded as \colorbox{blue!35}{Best}, \colorbox{blue!15}{Second best}, \colorbox{red!35}{Worst}, \colorbox{red!15}{Second worst}.}
\label{tab:in-species-test-set}
\end{table*}



\begin{table*}[h]
\centering
\renewcommand{\arraystretch}{0.9}
\resizebox{\textwidth}{!}{
\begin{tabular}{lccccccccccccc}
\toprule

\multirow{2}{*}{{Model}}     & \multicolumn{5}{c}{{Average Precision}}                                                                                 & \multicolumn{4}{c}{{F1@0.5}}                                                                    & \multicolumn{4}{c}{{F1@optimal threshold}}                                                      \\ 
\cmidrule(lr){2-6} \cmidrule(lr){7-10} \cmidrule(lr){11-14}
                             & mAP    & Adip  & Pelv  & Barb  & Dors  & Adip  & Pelv  & Barb  & Dors  & Adip     & Pelv     & Barb     & Dors    \\ 
 \midrule
VGG-19                     & 49.97                     & 68.42                     & 43.36                     & 65.47                     & 22.62                     & 79.93                     & 57.35                     & 79.55                     & 65.03                     & 81.18                     & 67.52                     & 81.05                     & 63.5                      \\
ResNet-18                  & 49.27                     & 81.46                     & 23.89                     & 80.68                     & 11.06                     & 79.16                     & 66.84                     & 83.77                     & 54.31                     & 85.16                     & 64.85                     & 84.0                      & 55.42                     \\
ResNet-34                  & 52.14                     & 79.8                      & 37.24                     & 75.45                     & 16.09                     & 86.67                     & 58.07                     & 82.87                     & 55.76                     & 86.67                     & 69.73                     & 82.87                     & 55.73                     \\
Inception-v3               & \colorbox{red!35}{28.93}  & 67.01                     & 6.33                      & \colorbox{red!35}{27.27}  & 15.13                     & 73.56                     & 55.49                     & \colorbox{red!35}{63.14}  & 57.37                     & 76.53                     & 49.61                     & \colorbox{red!35}{61.65}  & 61.51                     \\
ResNext-50                 & 53.0                      & 91.16                     & \colorbox{red!35}{1.42}   & 72.86                     & 46.56                     & 93.28                     & 48.29                     & 81.23                     & 73.4                      & 82.29                     & \colorbox{red!15}{48.71}  & 80.91                     & 72.85                     \\
MobileNet-v3              & 53.36                     & 89.93                     & 21.18                     & 60.63                     & 41.68                     & 89.58                     & 60.13                     & 77.15                     & 71.08                     & 90.16                     & 61.08                     & 74.65                     & 73.2                      \\
RegNet-y                   & 34.67                     & 67.04                     & 12.23                     & \colorbox{red!15}{45.17}  & 14.24                     & 80.03                     & 54.01                     & \colorbox{red!15}{69.27}  & 57.17                     & 75.4                      & 56.86                     & 71.16                     & 56.88                     \\
EfficientNet-v2            & 83.86                     & 95.45                     & 88.1                      & 83.97                     & 67.91                     & 95.15                     & 84.59                     & 86.76                     & \colorbox{blue!35}{79.31} & 92.76                     & 87.07                     & 82.62                     & \colorbox{blue!15}{79.43} \\
ConvNext-Base              & 79.6                      & 96.87                     & 73.18                     & 85.14                     & 63.21                     & 96.24                     & 84.32                     & 88.3                      & 74.24                     & 96.24                     & 77.98                     & 88.3                      & 78.71                     \\
\midrule
ViT-B-16                   & 47.27                     & 58.54                     & 40.13                     & 67.6                      & 22.82                     & \colorbox{red!15}{69.78}  & 66.66                     & 70.2                      & 62.41                     & \colorbox{red!15}{69.27}  & 65.89                     & 76.86                     & 61.87                     \\
ViT-B-32                   & 35.29                     & \colorbox{red!35}{47.21}  & 14.49                     & 50.29                     & 29.17                     & \colorbox{red!35}{69.3}   & 56.96                     & 69.79                     & 64.87                     & \colorbox{red!35}{62.59}  & 56.1                      & \colorbox{red!15}{64.72}  & 63.66                     \\
DEiT-distilled             & 61.74                     & 65.19                     & 43.15                     & 82.55                     & 56.07                     & 80.14                     & 63.26                     & 85.71                     & 76.08                     & 73.5                      & 70.35                     & 86.78                     & 75.87                     \\
SWIN-B                     & 60.22                     & 81.1                      & 38.66                     & 79.96                     & 41.15                     & 87.02                     & 71.63                     & 86.15                     & 69.24                     & 82.11                     & 59.69                     & 86.87                     & 68.83                     \\
SWIN-B-22k                 & 68.18                     & 89.0                      & 59.52                     & 80.03                     & 44.18                     & 92.04                     & 75.96                     & 85.03                     & 57.88                     & 91.5                      & 74.77                     & 83.88                     & 58.25                     \\
CVT-13                     & \colorbox{red!15}{28.94}  & 55.04                     & \colorbox{red!15}{1.84}   & 48.15                     & \colorbox{red!15}{10.73}  & 77.28                     & \colorbox{red!15}{48.23}  & 72.95                     & \colorbox{red!15}{52.77}  & 69.61                     & \colorbox{red!35}{47.13}  & 72.95                     & 54.55                     \\
Mobile-ViT-xs              & 34.23                     & 55.65                     & 13.99                     & 58.08                     & \colorbox{red!35}{9.2}    & 77.36                     & 56.12                     & 75.46                     & \colorbox{red!35}{52.03}  & 76.35                     & 54.39                     & 75.17                     & \colorbox{red!35}{52.67}  \\
Mobile-ViT-v2              & 33.76                     & \colorbox{red!15}{52.62}  & 3.74                      & 64.05                     & 14.63                     & 73.59                     & \colorbox{red!35}{46.78}  & 79.37                     & 54.76                     & 70.19                     & 51.98                     & 77.4                      & \colorbox{red!15}{53.75}  \\
MaxViT-t                   & 75.42                     & 87.18                     & 81.3                      & 76.03                     & 57.15                     & 88.61                     & 69.04                     & 79.62                     & 75.05                     & 88.21                     & 83.98                     & 84.27                     & 75.63                     \\
PVT-v2                     & 60.72                     & 83.91                     & 9.26                      & 89.3                      & 60.42                     & 80.42                     & 56.72                     & 88.41                     & \colorbox{blue!15}{78.74} & 79.24                     & 49.62                     & 90.49                     & 76.61                     \\
\midrule
Q2L (R34-22k SH)           & 79.86                     & 92.93                     & 67.54                     & 89.14                     & \colorbox{blue!15}{69.82} & 88.54                     & 85.56                     & 86.14                     & 70.02                     & 92.79                     & 88.51                     & 90.09                     & 78.74                     \\
Q2L (R34-22k MH)           & 74.64                     & 92.5                      & 50.37                     & 85.21                     & \colorbox{blue!35}{70.47} & 91.04                     & 78.74                     & 86.84                     & 70.83                     & 93.48                     & 78.74                     & 89.0                      & \colorbox{blue!35}{82.07} \\
Q2L (SWIN-22k SH)          & \colorbox{blue!35}{88.41} & \colorbox{blue!15}{98.61} & \colorbox{blue!15}{93.06} & \colorbox{blue!15}{97.3}  & 64.65                     & \colorbox{blue!35}{97.84} & \colorbox{blue!15}{92.75} & \colorbox{blue!35}{96.08} & 74.5                      & \colorbox{blue!15}{97.42} & \colorbox{blue!15}{93.98} & \colorbox{blue!15}{95.22} & 75.43                     \\
\textbf{Q2L (SWIN-22k MH)} & \colorbox{blue!15}{88.23} & \colorbox{blue!35}{99.17} & \colorbox{blue!35}{96.39} & \colorbox{blue!35}{97.62} & 59.76                     & \colorbox{blue!15}{97.49} & \colorbox{blue!35}{99.04} & \colorbox{blue!15}{95.84} & 73.98                     & \colorbox{blue!35}{97.84} & \colorbox{blue!35}{98.12} & \colorbox{blue!35}{95.84} & 76.69 \\
\bottomrule
\end{tabular}%
}
\caption{Trait identification results on the \textbf{leave-out-species test set}. Results are color-coded as \colorbox{blue!35}{Best}, \colorbox{blue!15}{Second best}, \colorbox{red!35}{Worst}, \colorbox{red!15}{Second worst}.}
\label{tab:leave-species-out-test-set}
\end{table*}




\begin{table*}[]
\centering
\renewcommand{\arraystretch}{0.9}
\resizebox{\textwidth}{!}{
\begin{tabular}{lccccccccccccc}
\toprule

\multirow{2}{*}{{Model}}     & \multicolumn{5}{c}{{Average Precision}}                                                                                 & \multicolumn{4}{c}{{F1@0.5}}                                                                    & \multicolumn{4}{c}{{F1@optimal threshold}}                                                      \\ 
\cmidrule(lr){2-6} \cmidrule(lr){7-10} \cmidrule(lr){11-14}
                             & mAP    & Adip  & Pelv  & Barb  & Dors  & Adip  & Pelv  & Barb  & Dors  & Adip     & Pelv     & Barb     & Dors    \\ 
 \midrule
VGG-19            & 38.48                     & 49.36                     & 23.15                     & 35.79                     & \colorbox{red!15}{45.61}  & 64.89                     & 58.48                     & 62.53                     & 64.09                     & 70.12                     & 59.19                     & 66.48                     & 65.87                     \\
ResNet-18         & 39.61                     & 46.24                     & 18.42                     & 42.02                     & 51.74                     & 58.86                     & 57.21                     & 68.76                     & \colorbox{red!15}{59.42}  & 69.07                     & 57.69                     & 70.89                     & 66.68                     \\
ResNet-34         & 45.53                     & 58.78                     & 25.66                     & 38.66                     & 59.02                     & 75.17                     & 61.16                     & 66.45                     & 70.64                     & 74.81                     & 56.85                     & 66.9                      & 68.02                     \\
Inception-v3      & \colorbox{red!35}{30.07}  & 48.07                     & \colorbox{red!15}{14.22}  & \colorbox{red!35}{22.66}  & \colorbox{red!35}{35.32}  & 58.86                     & \colorbox{red!35}{49.07}  & \colorbox{red!35}{58.26}  & \colorbox{red!35}{53.36}  & 72.76                     & \colorbox{red!35}{48.09}  & \colorbox{red!15}{60.23}  & \colorbox{red!35}{55.66}  \\
ResNext-50        & 43.25                     & 57.02                     & 19.56                     & 36.62                     & 59.79                     & 70.12                     & \colorbox{red!15}{52.37}  & 63.25                     & 62.53                     & 73.58                     & 52.37                     & 62.67                     & 68.89                     \\
MobileNet-v3      & 41.99                     & 46.18                     & 27.96                     & 30.44                     & 63.36                     & 69.87                     & 57.63                     & 61.03                     & 70.68                     & 70.5                      & 56.75                     & 61.72                     & 71.85                     \\
RegNet-y          & 38.1                      & 43.45                     & 23.14                     & 31.93                     & 53.87                     & 66.74                     & 54.87                     & 64.19                     & 66.65                     & 70.37                     & \colorbox{red!15}{52.25}  & 66.74                     & 66.56                     \\
EfficientNet-v2   & 55.36                     & \colorbox{blue!35}{63.96} & 30.75                     & 55.08                     & 71.66                     & 74.03                     & 60.84                     & 73.13                     & 75.42                     & \colorbox{blue!35}{76.35} & 58.07                     & 75.28                     & 78.06                     \\
ConvNext-Base     & 53.96                     & 61.27                     & 38.19                     & 47.55                     & 68.84                     & 73.39                     & 69.59                     & 66.03                     & 78.68                     & 73.5                      & 67.52                     & 68.63                     & 77.02                     \\
\midrule
ViT-B-16          & 37.63                     & 37.69                     & 26.92                     & 36.72                     & 49.2                      & 64.72                     & 58.76                     & 66.35                     & 66.57                     & 65.74                     & 60.47                     & 66.61                     & 67.57                     \\
ViT-B-32          & 33.7                      & \colorbox{red!35}{33.43}  & 24.34                     & 30.47                     & 46.54                     & \colorbox{red!35}{58.11}  & 60.19                     & 60.33                     & 63.91                     & \colorbox{red!15}{62.28}  & 56.96                     & 63.14                     & \colorbox{red!15}{65.76}  \\
DEiT-distilled    & 40.69                     & 40.36                     & 32.82                     & 29.5                      & 60.08                     & 61.85                     & 64.24                     & \colorbox{red!15}{58.77}  & 69.82                     & 67.34                     & 61.97                     & 64.11                     & 70.63                     \\
SWIN-B            & 44.68                     & 40.67                     & 35.6                      & 36.36                     & 66.07                     & 64.16                     & 62.32                     & 67.73                     & 74.7                      & 68.0                      & 60.16                     & 68.14                     & 76.93                     \\
SWIN-B-22k        & 51.53                     & 55.55                     & 32.2                      & 50.54                     & 67.82                     & 72.51                     & 65.99                     & 73.78                     & 75.45                     & 72.21                     & 63.0                      & 71.46                     & 74.37                     \\
CVT-13            & 34.76                     & \colorbox{red!15}{35.87}  & 24.28                     & 30.76                     & 48.13                     & \colorbox{red!15}{58.53}  & 53.3                      & 59.69                     & 63.48                     & \colorbox{red!35}{61.84}  & 57.93                     & \colorbox{red!35}{59.69}  & 67.91                     \\
Mobile-ViT-xs     & \colorbox{red!15}{30.6}   & 39.08                     & \colorbox{red!35}{13.7}   & \colorbox{red!15}{23.99}  & 45.62                     & 65.27                     & 54.82                     & 62.68                     & 64.64                     & 69.42                     & 52.95                     & 61.58                     & 67.46                     \\
Mobile-ViT-v2     & 34.89                     & 44.5                      & 19.84                     & 28.45                     & 46.78                     & 67.61                     & 57.85                     & 63.99                     & 63.17                     & 68.33                     & 54.95                     & 65.63                     & 65.83                     \\
MaxViT-t          & 49.67                     & 50.34                     & 33.71                     & 51.55                     & 63.07                     & 69.21                     & 65.67                     & 74.5                      & 72.52                     & 68.97                     & 63.33                     & 72.03                     & 71.53                     \\
PVT-v2            & 46.42                     & 44.53                     & 27.7                      & 47.75                     & 65.71                     & 62.68                     & 62.42                     & 71.54                     & 73.56                     & 64.08                     & 56.96                     & 71.45                     & 75.24                     \\
\midrule
Q2L (R34-22k SH)  & 55.84                     & \colorbox{blue!15}{61.67} & 36.8                      & 50.44                     & \colorbox{blue!35}{74.43} & \colorbox{blue!35}{76.16} & 67.35                     & 73.61                     & \colorbox{blue!35}{80.07} & \colorbox{blue!15}{75.23} & 69.25                     & 72.76                     & \colorbox{blue!35}{79.63} \\
Q2L (R34-22k MH)  & 53.98                     & 56.6                      & 40.04                     & 45.91                     & \colorbox{blue!15}{73.36} & 75.42                     & 67.54                     & 72.57                     & \colorbox{blue!15}{78.9}  & 70.93                     & 67.88                     & 72.18                     & 77.8                      \\
Q2L (SWIN-22k SH) & \colorbox{blue!35}{59.39} & 59.18                     & \colorbox{blue!15}{41.65} & \colorbox{blue!35}{63.46} & 73.28                     & \colorbox{blue!15}{75.74} & \colorbox{blue!35}{70.83} & \colorbox{blue!35}{79.16} & 78.69                     & 74.8                      & \colorbox{blue!15}{70.54} & \colorbox{blue!35}{77.78} & \colorbox{blue!15}{78.44} \\
Q2L (SWIN-22k MH) & \colorbox{blue!15}{58.27} & 57.97                     & \colorbox{blue!35}{42.71} & \colorbox{blue!15}{60.56} & 71.83                     & 74.55                     & \colorbox{blue!35}{70.83} & \colorbox{blue!15}{77.89} & 78.18                     & 73.52                     & \colorbox{blue!35}{71.46} & \colorbox{blue!15}{77.68} & 75.53  \\
\bottomrule
\end{tabular}%
}
\caption{Trait identification results on the challenging \textbf{manual-annotation test set}. All models struggle to identify traits on the diverse set of species contained within the manual-annotation set. Results are color-coded as \colorbox{blue!35}{Best}, \colorbox{blue!15}{Second best}, \colorbox{red!35}{Worst}, \colorbox{red!15}{Second worst}.
}
\label{tab:manually-annotated-test-set}
\end{table*}






%% file: main.bbl
\begin{thebibliography}{78}
\providecommand{\natexlab}[1]{#1}
\providecommand{\url}[1]{\texttt{#1}}
\expandafter\ifx\csname urlstyle\endcsname\relax
  \providecommand{\doi}[1]{doi: #1}\else
  \providecommand{\doi}{doi: \begingroup \urlstyle{rm}\Url}\fi

\bibitem[Mor()]{Morphbank}
Morphbank: Biological imaging (https://www.morphbank.net/).
\newblock \emph{Florida State University, Department of Scientific Computing, Tallahassee, FL 32306-4026 USA.}

\bibitem[fis()]{fishair}
Multimedia of fish specimen and associated metadata. fish-air.
\newblock \emph{Biology guided Neural Network. Tulane University Biodiversity Research Institute (https://fishair.org).}

\bibitem[fmn()]{fmnh}
Fmnh field museum of natural history (zoology) fish collection.
\newblock \emph{Field Museum. https://fmipt.fieldmuseum.org/ipt/resource?r=fmnh\_fishes}.

\bibitem[gli()]{glin}
Great lakes invasives network project.
\newblock \emph{https://greatlakesinvasives.org/portal/index.php}.

\bibitem[uma()]{umadison}
University of wisconsin-madison zoological museum - fish.
\newblock \emph{http://zoology.wisc.edu/uwzm/}.

\bibitem[umm()]{ummz}
Ummz university of michigan museum of zoology, division of fishes.
\newblock \emph{https://ipt.lsa.umich.edu/resource?r=ummz\_fish}.

\bibitem[IDi(2020)]{IDigBio}
idigbio.
\newblock \emph{http://www.idigbio.org/portal}, 2020.

\bibitem[inh(2022)]{inhs}
Inhs collections data.
\newblock \emph{http://biocoll.inhs.illinois.edu/portal/index.php}, 2022.

\bibitem[jfb(2022)]{jfbm}
Jfbm bell atlas.
\newblock \emph{http://bellatlas.umn.edu/index.php.}, 2022.

\bibitem[Anantharajah et~al.(2014)Anantharajah, Ge, McCool, Denman, Fookes, Corke, Tjondronegoro, and Sridharan]{anantharajah2014local}
Kaneswaran Anantharajah, ZongYuan Ge, Christopher McCool, Simon Denman, Clinton~B Fookes, Peter Corke, Dian~W Tjondronegoro, and Sridha Sridharan.
\newblock Local inter-session variability modelling for object classification.
\newblock In \emph{Winter Conference on Applications of Computer Vision (WACV), 2013 IEEE Conference on}, 2014.

\bibitem[Chen et~al.(2017)Chen, Papandreou, Schroff, and Adam]{chen2017rethinking}
Liang-Chieh Chen, George Papandreou, Florian Schroff, and Hartwig Adam.
\newblock Rethinking atrous convolution for semantic image segmentation.
\newblock \emph{arXiv preprint arXiv:1706.05587}, 2017.

\bibitem[Chen et~al.(2018)Chen, Zhu, Papandreou, Schroff, and Adam]{deeplabv3plus2018}
Liang-Chieh Chen, Yukun Zhu, George Papandreou, Florian Schroff, and Hartwig Adam.
\newblock Encoder-decoder with atrous separable convolution for semantic image segmentation.
\newblock In \emph{ECCV}, 2018.

\bibitem[Cheng et~al.(2021)Cheng, Schwing, and Kirillov]{cheng2021per}
Bowen Cheng, Alex Schwing, and Alexander Kirillov.
\newblock Per-pixel classification is not all you need for semantic segmentation.
\newblock \emph{Advances in neural information processing systems}, 34:\penalty0 17864--17875, 2021.

\bibitem[Cui et~al.(2019)Cui, Jia, Lin, Song, and Belongie]{cui2019class}
Yin Cui, Menglin Jia, Tsung-Yi Lin, Yang Song, and Serge Belongie.
\newblock Class-balanced loss based on effective number of samples.
\newblock In \emph{Proceedings of the IEEE/CVF conference on computer vision and pattern recognition}, pages 9268--9277, 2019.

\bibitem[{CVAT.ai Corporation}(2023)]{cvat2023}
{CVAT.ai Corporation}.
\newblock Computer vision annotation tool (cvat) (v2.4.3), 2023.

\bibitem[Daly~M(2018)]{osum}
Johnson~N Daly~M.
\newblock Ohio state university fish division (osum).
\newblock \emph{Museum of Biological Diversity, The Ohio State University. Occurrence dataset, https://doi.org/10.15468/subsl8}, 2018.

\bibitem[Deitke et~al.(2024)Deitke, Clark, Lee, Tripathi, Yang, Park, Salehi, Muennighoff, Lo, Soldaini, et~al.]{deitke2024molmo}
Matt Deitke, Christopher Clark, Sangho Lee, Rohun Tripathi, Yue Yang, Jae~Sung Park, Mohammadreza Salehi, Niklas Muennighoff, Kyle Lo, Luca Soldaini, et~al.
\newblock Molmo and pixmo: Open weights and open data for state-of-the-art multimodal models.
\newblock \emph{arXiv preprint arXiv:2409.17146}, 2024.

\bibitem[Deng et~al.(2009{\natexlab{a}})Deng, Dong, Socher, Li, Li, and Fei-Fei]{deng2009imagenet}
Jia Deng, Wei Dong, Richard Socher, Li-Jia Li, Kai Li, and Li Fei-Fei.
\newblock Imagenet: A large-scale hierarchical image database.
\newblock In \emph{2009 IEEE conference on computer vision and pattern recognition}, pages 248--255. Ieee, 2009{\natexlab{a}}.

\bibitem[Deng et~al.(2009{\natexlab{b}})Deng, Dong, Socher, Li, Li, and Fei-Fei]{imagenet}
Jia Deng, Wei Dong, Richard Socher, Li-Jia Li, Kai Li, and Li Fei-Fei.
\newblock Imagenet: A large-scale hierarchical image database.
\newblock In \emph{2009 IEEE Conference on Computer Vision and Pattern Recognition}, pages 248--255, 2009{\natexlab{b}}.

\bibitem[Dosovitskiy et~al.(2020)Dosovitskiy, Beyer, Kolesnikov, Weissenborn, Zhai, Unterthiner, Dehghani, Minderer, Heigold, Gelly, et~al.]{ViT}
Alexey Dosovitskiy, Lucas Beyer, Alexander Kolesnikov, Dirk Weissenborn, Xiaohua Zhai, Thomas Unterthiner, Mostafa Dehghani, Matthias Minderer, Georg Heigold, Sylvain Gelly, et~al.
\newblock An image is worth 16x16 words: Transformers for image recognition at scale.
\newblock \emph{arXiv preprint arXiv:2010.11929}, 2020.

\bibitem[Elhamod et~al.(2023)Elhamod, Khurana, Manogaran, Uyeda, Balk, Dahdul, Bakis, Bart~Jr, Mabee, Lapp, et~al.]{elhamod2023discovering}
Mohannad Elhamod, Mridul Khurana, Harish~Babu Manogaran, Josef~C Uyeda, Meghan~A Balk, Wasila Dahdul, Yasin Bakis, Henry~L Bart~Jr, Paula~M Mabee, Hilmar Lapp, et~al.
\newblock Discovering novel biological traits from images using phylogeny-guided neural networks.
\newblock In \emph{Proceedings of the 29th ACM SIGKDD Conference on Knowledge Discovery and Data Mining}, pages 3966--3978, 2023.

\bibitem[Fergus et~al.(2024)Fergus, Chalmers, Longmore, and Wich]{fergus2024harnessing}
Paul Fergus, Carl Chalmers, Steven Longmore, and Serge Wich.
\newblock Harnessing artificial intelligence for wildlife conservation.
\newblock \emph{Conservation}, 4\penalty0 (4):\penalty0 685--702, 2024.

\bibitem[Fisher et~al.(2016)Fisher, Chen{-}Burger, Giordano, Hardman, and Lin]{DBLP:series/isrl/104}
Robert~B. Fisher, Yun{-}Heh Chen{-}Burger, Daniela Giordano, Lynda Hardman, and Fang{-}Pang Lin, editors.
\newblock \emph{Fish4Knowledge: Collecting and Analyzing Massive Coral Reef Fish Video Data}.
\newblock Springer, 2016.

\bibitem[Froese and Pauly(2024)]{fishbase}
R. Froese and D. Pauly.
\newblock Fishbase, 2024.
\newblock World Wide Web electronic publication. Version 02/2024.

\bibitem[Gharaee et~al.(2024{\natexlab{a}})Gharaee, Gong, Pellegrino, Zarubiieva, Haurum, Lowe, McKeown, Ho, McLeod, Wei, et~al.]{gharaee2024step}
Zahra Gharaee, ZeMing Gong, Nicholas Pellegrino, Iuliia Zarubiieva, Joakim~Bruslund Haurum, Scott Lowe, Jaclyn McKeown, Chris Ho, Joschka McLeod, Yi-Yun Wei, et~al.
\newblock A step towards worldwide biodiversity assessment: The bioscan-1m insect dataset.
\newblock \emph{Advances in Neural Information Processing Systems}, 36, 2024{\natexlab{a}}.

\bibitem[Gharaee et~al.(2024{\natexlab{b}})Gharaee, Lowe, Gong, Arias, Pellegrino, Wang, Haurum, Zarubiieva, Kari, Steinke, et~al.]{gharaee2024bioscan}
Zahra Gharaee, Scott~C Lowe, ZeMing Gong, Pablo~Millan Arias, Nicholas Pellegrino, Austin~T Wang, Joakim~Bruslund Haurum, Iuliia Zarubiieva, Lila Kari, Dirk Steinke, et~al.
\newblock Bioscan-5m: A multimodal dataset for insect biodiversity.
\newblock \emph{arXiv preprint arXiv:2406.12723}, 2024{\natexlab{b}}.

\bibitem[Graham and Harrod(2009)]{graham2009implications}
CT Graham and Chris Harrod.
\newblock Implications of climate change for the fishes of the british isles.
\newblock \emph{Journal of Fish Biology}, 74\penalty0 (6):\penalty0 1143--1205, 2009.

\bibitem[He et~al.(2022)He, Chen, Liu, Kortylewski, Yang, Bai, and Wang]{he2022transfg}
Ju He, Jie-Neng Chen, Shuai Liu, Adam Kortylewski, Cheng Yang, Yutong Bai, and Changhu Wang.
\newblock Transfg: A transformer architecture for fine-grained recognition.
\newblock In \emph{Proceedings of the AAAI conference on artificial intelligence}, pages 852--860, 2022.

\bibitem[He et~al.(2016)He, Zhang, Ren, and Sun]{resnet}
Kaiming He, Xiangyu Zhang, Shaoqing Ren, and Jian Sun.
\newblock Deep residual learning for image recognition.
\newblock In \emph{Proceedings of the IEEE conference on computer vision and pattern recognition}, pages 770--778, 2016.

\bibitem[Houle and Rossoni(2022{\natexlab{a}})]{houle2022complexity}
David Houle and Daniela~M Rossoni.
\newblock Complexity, evolvability, and the process of adaptation.
\newblock \emph{Annual Review of Ecology, Evolution, and Systematics}, 53:\penalty0 137--159, 2022{\natexlab{a}}.

\bibitem[Houle and Rossoni(2022{\natexlab{b}})]{houlerossoni2022}
David Houle and Daniela~M Rossoni.
\newblock Complexity, evolvability, and the process of adaptation.
\newblock \emph{Annual Review of Ecology, Evolution, and Systematics}, 53, 2022{\natexlab{b}}.

\bibitem[Howard et~al.(2019)Howard, Sandler, Chu, Chen, Chen, Tan, Wang, Zhu, Pang, Vasudevan, et~al.]{howard2019searching}
Andrew Howard, Mark Sandler, Grace Chu, Liang-Chieh Chen, Bo Chen, Mingxing Tan, Weijun Wang, Yukun Zhu, Ruoming Pang, Vijay Vasudevan, et~al.
\newblock Searching for mobilenetv3.
\newblock In \emph{Proceedings of the IEEE/CVF international conference on computer vision}, pages 1314--1324, 2019.

\bibitem[Katija et~al.(2022)Katija, Orenstein, Schlining, Lundsten, Barnard, Sainz, Boulais, Cromwell, Butler, Woodward, et~al.]{katija2022fathomnet}
Kakani Katija, Eric Orenstein, Brian Schlining, Lonny Lundsten, Kevin Barnard, Giovanna Sainz, Oceane Boulais, Megan Cromwell, Erin Butler, Benjamin Woodward, et~al.
\newblock Fathomnet: A global image database for enabling artificial intelligence in the ocean.
\newblock \emph{Scientific reports}, 12\penalty0 (1):\penalty0 15914, 2022.

\bibitem[Khan et~al.(2023)Khan, Li, Temple, and Elhoseiny]{khan2023fishnet}
Faizan~Farooq Khan, Xiang Li, Andrew~J Temple, and Mohamed Elhoseiny.
\newblock Fishnet: A large-scale dataset and benchmark for fish recognition, detection, and functional trait prediction.
\newblock In \emph{Proceedings of the IEEE/CVF International Conference on Computer Vision}, pages 20496--20506, 2023.

\bibitem[Khosla et~al.(2011)Khosla, Jayadevaprakash, Yao, and Li]{khosla2011novel}
Aditya Khosla, Nityananda Jayadevaprakash, Bangpeng Yao, and Fei-Fei Li.
\newblock Novel dataset for fine-grained image categorization: Stanford dogs.
\newblock In \emph{Proceedings CVPR workshop on fine-grained visual categorization (FGVC)}, 2011.

\bibitem[Kim et~al.(2020)Kim, Jeong, and Shin]{kim2020m2m}
Jaehyung Kim, Jongheon Jeong, and Jinwoo Shin.
\newblock M2m: Imbalanced classification via major-to-minor translation.
\newblock In \emph{Proceedings of the IEEE/CVF conference on computer vision and pattern recognition}, pages 13896--13905, 2020.

\bibitem[Kirillov et~al.(2019)Kirillov, Girshick, He, and Doll{\'a}r]{kirillov2019panoptic}
Alexander Kirillov, Ross Girshick, Kaiming He, and Piotr Doll{\'a}r.
\newblock Panoptic feature pyramid networks.
\newblock In \emph{Proceedings of the IEEE/CVF Conference on Computer Vision and Pattern Recognition}, pages 6399--6408, 2019.

\bibitem[Kirillov et~al.(2023)Kirillov, Mintun, Ravi, Mao, Rolland, Gustafson, Xiao, Whitehead, Berg, Lo, et~al.]{kirillov2023segment}
Alexander Kirillov, Eric Mintun, Nikhila Ravi, Hanzi Mao, Chloe Rolland, Laura Gustafson, Tete Xiao, Spencer Whitehead, Alexander~C Berg, Wan-Yen Lo, et~al.
\newblock Segment anything.
\newblock In \emph{Proceedings of the IEEE/CVF International Conference on Computer Vision}, pages 4015--4026, 2023.

\bibitem[Lin et~al.(2017)Lin, Goyal, Girshick, He, and Doll{\'a}r]{lin2017focal}
Tsung-Yi Lin, Priya Goyal, Ross Girshick, Kaiming He, and Piotr Doll{\'a}r.
\newblock Focal loss for dense object detection.
\newblock In \emph{Proceedings of the IEEE international conference on computer vision}, pages 2980--2988, 2017.

\bibitem[Liu et~al.(2021{\natexlab{a}})Liu, Zhang, Yang, Su, and Zhu]{liu2021query2label}
Shilong Liu, Lei Zhang, Xiao Yang, Hang Su, and Jun Zhu.
\newblock Query2label: A simple transformer way to multi-label classification.
\newblock \emph{arXiv preprint arXiv:2107.10834}, 2021{\natexlab{a}}.

\bibitem[Liu et~al.(2023)Liu, Zeng, Ren, Li, Zhang, Yang, Li, Yang, Su, Zhu, et~al.]{liu2023grounding}
Shilong Liu, Zhaoyang Zeng, Tianhe Ren, Feng Li, Hao Zhang, Jie Yang, Chunyuan Li, Jianwei Yang, Hang Su, Jun Zhu, et~al.
\newblock Grounding dino: Marrying dino with grounded pre-training for open-set object detection.
\newblock \emph{arXiv preprint arXiv:2303.05499}, 2023.

\bibitem[Liu et~al.(2021{\natexlab{b}})Liu, Lin, Cao, Hu, Wei, Zhang, Lin, and Guo]{swin}
Ze Liu, Yutong Lin, Yue Cao, Han Hu, Yixuan Wei, Zheng Zhang, Stephen Lin, and Baining Guo.
\newblock Swin transformer: Hierarchical vision transformer using shifted windows.
\newblock In \emph{Proceedings of the IEEE/CVF international conference on computer vision}, pages 10012--10022, 2021{\natexlab{b}}.

\bibitem[Liu et~al.(2022)Liu, Mao, Wu, Feichtenhofer, Darrell, and Xie]{convnext}
Zhuang Liu, Hanzi Mao, Chao-Yuan Wu, Christoph Feichtenhofer, Trevor Darrell, and Saining Xie.
\newblock A convnet for the 2020s.
\newblock In \emph{Proceedings of the IEEE/CVF conference on computer vision and pattern recognition}, pages 11976--11986, 2022.

\bibitem[Mabee et~al.(2018)Mabee, Dahdul, Balhoff, Lapp, Manda, Uyeda, Vision, and Westerfield]{mabee2018phenoscape}
Paula~M Mabee, Wasila~M Dahdul, James~P Balhoff, Hilmar Lapp, Prashanti Manda, Josef Uyeda, Todd Vision, and Monte Westerfield.
\newblock Phenoscape: semantic analysis of organismal traits and genes yields insights in evolutionary biology.
\newblock In \emph{Application of Semantic Technology in Biodiversity Science}, pages 207--224. IOS Press, 2018.

\bibitem[Mehta and Rastegari(2021)]{mehta2021mobilevit}
Sachin Mehta and Mohammad Rastegari.
\newblock Mobilevit: light-weight, general-purpose, and mobile-friendly vision transformer.
\newblock \emph{arXiv preprint arXiv:2110.02178}, 2021.

\bibitem[of~Life et~al.(2019)of~Life, Cranston, Redelings, Reyes, Allman, McTavish, and Holder]{opentreeoflife2019}
Open~Tree of Life, Karen~A. Cranston, Benjamin Redelings, Luna Luisa~Sanchez Reyes, Jim Allman, Emily~Jane McTavish, and Mark~T. Holder.
\newblock {Open Tree of Life Taxonomy (3.2)}.
\newblock Zenodo, 2019.

\bibitem[Oquab et~al.(2023)Oquab, Darcet, Moutakanni, Vo, Szafraniec, Khalidov, Fernandez, Haziza, Massa, El-Nouby, et~al.]{oquab2023dinov2}
Maxime Oquab, Timoth{\'e}e Darcet, Th{\'e}o Moutakanni, Huy Vo, Marc Szafraniec, Vasil Khalidov, Pierre Fernandez, Daniel Haziza, Francisco Massa, Alaaeldin El-Nouby, et~al.
\newblock Dinov2: Learning robust visual features without supervision.
\newblock \emph{arXiv preprint arXiv:2304.07193}, 2023.

\bibitem[Parkhi et~al.(2012)Parkhi, Vedaldi, Zisserman, and Jawahar]{parkhi2012cats}
Omkar~M Parkhi, Andrea Vedaldi, Andrew Zisserman, and CV Jawahar.
\newblock Cats and dogs.
\newblock In \emph{2012 IEEE conference on computer vision and pattern recognition}, pages 3498--3505, 2012.

\bibitem[Paul et~al.(2023)Paul, Chowdhury, Xiong, Chang, Carlyn, Stevens, Provost, Karpatne, Carstens, Rubenstein, et~al.]{paul2023simple}
Dipanjyoti Paul, Arpita Chowdhury, Xinqi Xiong, Feng-Ju Chang, David Carlyn, Samuel Stevens, Kaiya~L Provost, Anuj Karpatne, Bryan Carstens, Daniel Rubenstein, et~al.
\newblock A simple interpretable transformer for fine-grained image classification and analysis.
\newblock \emph{arXiv preprint arXiv:2311.04157}, 2023.

\bibitem[Pepper et~al.(2021)Pepper, Greenberg, Baki{\c{s}}, Wang, Bart, and Breen]{pepper2021automatic}
Joel Pepper, Jane Greenberg, Yasin Baki{\c{s}}, Xiaojun Wang, Henry Bart, and David Breen.
\newblock Automatic metadata generation for fish specimen image collections.
\newblock In \emph{2021 ACM/IEEE Joint Conference on Digital Libraries (JCDL)}, pages 31--40. IEEE, 2021.

\bibitem[Piosenka(2023)]{piosenka2023birds}
Gerald Piosenka.
\newblock Birds 525 species - image classification.
\newblock 2023.

\bibitem[Price et~al.(2022)Price, Friedman, Corn, Larouche, Brockelsby, Lee, Nagaraj, Bertrand, Danao, Coyne, et~al.]{price2022fishshapes}
Samantha~A Price, Sarah~T Friedman, Katherine~A Corn, Olivier Larouche, Kasey Brockelsby, Anna~J Lee, Maya Nagaraj, Nick~G Bertrand, Mailee Danao, Megan~C Coyne, et~al.
\newblock Fishshapes v1: Functionally relevant measurements of teleost shape and size on three dimensions, 2022.

\bibitem[Radford et~al.(2021{\natexlab{a}})Radford, Kim, Hallacy, Ramesh, Goh, Agarwal, Sastry, Askell, Mishkin, Clark, et~al.]{clip}
Alec Radford, Jong~Wook Kim, Chris Hallacy, Aditya Ramesh, Gabriel Goh, Sandhini Agarwal, Girish Sastry, Amanda Askell, Pamela Mishkin, Jack Clark, et~al.
\newblock Learning transferable visual models from natural language supervision.
\newblock In \emph{International conference on machine learning}, pages 8748--8763. PMLR, 2021{\natexlab{a}}.

\bibitem[Radford et~al.(2021{\natexlab{b}})Radford, Kim, Hallacy, Ramesh, Goh, Agarwal, Sastry, Askell, Mishkin, Clark, et~al.]{radford2021learning}
Alec Radford, Jong~Wook Kim, Chris Hallacy, Aditya Ramesh, Gabriel Goh, Sandhini Agarwal, Girish Sastry, Amanda Askell, Pamela Mishkin, Jack Clark, et~al.
\newblock Learning transferable visual models from natural language supervision.
\newblock In \emph{International conference on machine learning}, pages 8748--8763. PMLR, 2021{\natexlab{b}}.

\bibitem[Radosavovic et~al.(2020)Radosavovic, Kosaraju, Girshick, He, and Doll{\'a}r]{radosavovic2020designing}
Ilija Radosavovic, Raj~Prateek Kosaraju, Ross Girshick, Kaiming He, and Piotr Doll{\'a}r.
\newblock Designing network design spaces.
\newblock In \emph{Proceedings of the IEEE/CVF conference on computer vision and pattern recognition}, pages 10428--10436, 2020.

\bibitem[Ravi et~al.(2024)Ravi, Gabeur, Hu, Hu, Ryali, Ma, Khedr, R{\"a}dle, Rolland, Gustafson, et~al.]{ravi2024sam}
Nikhila Ravi, Valentin Gabeur, Yuan-Ting Hu, Ronghang Hu, Chaitanya Ryali, Tengyu Ma, Haitham Khedr, Roman R{\"a}dle, Chloe Rolland, Laura Gustafson, et~al.
\newblock Sam 2: Segment anything in images and videos.
\newblock \emph{arXiv preprint arXiv:2408.00714}, 2024.

\bibitem[Ronneberger et~al.(2015)Ronneberger, Fischer, and Brox]{ronneberger2015u}
Olaf Ronneberger, Philipp Fischer, and Thomas Brox.
\newblock U-net: Convolutional networks for biomedical image segmentation.
\newblock In \emph{International Conference on Medical image computing and computer-assisted intervention}, pages 234--241. Springer, 2015.

\bibitem[Saleh et~al.(2020)Saleh, Laradji, Konovalov, Bradley, Vazquez, and Sheaves]{saleh2020realistic}
Alzayat Saleh, Issam~H Laradji, Dmitry~A Konovalov, Michael Bradley, David Vazquez, and Marcus Sheaves.
\newblock A realistic fish-habitat dataset to evaluate algorithms for underwater visual analysis.
\newblock \emph{Scientific Reports}, 10\penalty0 (1):\penalty0 14671, 2020.

\bibitem[Shwartz-Ziv et~al.(2024)Shwartz-Ziv, Goldblum, Li, Bruss, and Wilson]{shwartz2024simplifying}
Ravid Shwartz-Ziv, Micah Goldblum, Yucen Li, C~Bayan Bruss, and Andrew~G Wilson.
\newblock Simplifying neural network training under class imbalance.
\newblock \emph{Advances in Neural Information Processing Systems}, 36, 2024.

\bibitem[Simonyan and Zisserman(2014)]{simonyan2014very}
Karen Simonyan and Andrew Zisserman.
\newblock Very deep convolutional networks for large-scale image recognition.
\newblock \emph{arXiv preprint arXiv:1409.1556}, 2014.

\bibitem[Steiner et~al.(2021)Steiner, Kolesnikov, Zhai, Wightman, Uszkoreit, and Beyer]{steiner2021train}
Andreas Steiner, Alexander Kolesnikov, Xiaohua Zhai, Ross Wightman, Jakob Uszkoreit, and Lucas Beyer.
\newblock How to train your vit? data, augmentation, and regularization in vision transformers.
\newblock \emph{arXiv preprint arXiv:2106.10270}, 2021.

\bibitem[Stevens et~al.(2023)Stevens, Wu, Thompson, Campolongo, Song, Carlyn, Dong, Dahdul, Stewart, Berger-Wolf, et~al.]{bioclip}
Samuel Stevens, Jiaman Wu, Matthew~J Thompson, Elizabeth~G Campolongo, Chan~Hee Song, David~Edward Carlyn, Li Dong, Wasila~M Dahdul, Charles Stewart, Tanya Berger-Wolf, et~al.
\newblock Bioclip: A vision foundation model for the tree of life.
\newblock \emph{arXiv preprint arXiv:2311.18803}, 2023.

\bibitem[Stevens et~al.(2024)Stevens, Wu, Thompson, Campolongo, Song, Carlyn, Dong, Dahdul, Stewart, Berger-Wolf, et~al.]{stevens2024bioclip}
Samuel Stevens, Jiaman Wu, Matthew~J Thompson, Elizabeth~G Campolongo, Chan~Hee Song, David~Edward Carlyn, Li Dong, Wasila~M Dahdul, Charles Stewart, Tanya Berger-Wolf, et~al.
\newblock Bioclip: A vision foundation model for the tree of life.
\newblock In \emph{Proceedings of the IEEE/CVF Conference on Computer Vision and Pattern Recognition}, pages 19412--19424, 2024.

\bibitem[Szegedy et~al.(2016)Szegedy, Vanhoucke, Ioffe, Shlens, and Wojna]{inceptionv3}
Christian Szegedy, Vincent Vanhoucke, Sergey Ioffe, Jon Shlens, and Zbigniew Wojna.
\newblock Rethinking the inception architecture for computer vision.
\newblock In \emph{Proceedings of the IEEE conference on computer vision and pattern recognition}, pages 2818--2826, 2016.

\bibitem[Tan and Le(2021)]{tan2021efficientnetv2}
Mingxing Tan and Quoc Le.
\newblock Efficientnetv2: Smaller models and faster training.
\newblock In \emph{International conference on machine learning}, pages 10096--10106. PMLR, 2021.

\bibitem[Touvron et~al.(2021)Touvron, Cord, Douze, Massa, Sablayrolles, and J{\'e}gou]{touvron2021training}
Hugo Touvron, Matthieu Cord, Matthijs Douze, Francisco Massa, Alexandre Sablayrolles, and Herv{\'e} J{\'e}gou.
\newblock Training data-efficient image transformers \& distillation through attention.
\newblock In \emph{International conference on machine learning}, pages 10347--10357. PMLR, 2021.

\bibitem[Tu et~al.(2022)Tu, Talebi, Zhang, Yang, Milanfar, Bovik, and Li]{maxvit}
Zhengzhong Tu, Hossein Talebi, Han Zhang, Feng Yang, Peyman Milanfar, Alan Bovik, and Yinxiao Li.
\newblock Maxvit: Multi-axis vision transformer.
\newblock In \emph{European conference on computer vision}, pages 459--479. Springer, 2022.

\bibitem[Tuia et~al.(2022)Tuia, Kellenberger, Beery, Costelloe, Zuffi, Risse, Mathis, Mathis, Van~Langevelde, Burghardt, et~al.]{tuia2022perspectives}
Devis Tuia, Benjamin Kellenberger, Sara Beery, Blair~R Costelloe, Silvia Zuffi, Benjamin Risse, Alexander Mathis, Mackenzie~W Mathis, Frank Van~Langevelde, Tilo Burghardt, et~al.
\newblock Perspectives in machine learning for wildlife conservation.
\newblock \emph{Nature communications}, 13\penalty0 (1):\penalty0 1--15, 2022.

\bibitem[Ulucan et~al.(2020)Ulucan, Karakaya, and Turkan]{ulucan2020large}
Oguzhan Ulucan, Diclehan Karakaya, and Mehmet Turkan.
\newblock A large-scale dataset for fish segmentation and classification.
\newblock In \emph{2020 Innovations in Intelligent Systems and Applications Conference (ASYU)}, pages 1--5. IEEE, 2020.

\bibitem[Van~Horn et~al.(2015)Van~Horn, Branson, Farrell, Haber, Barry, Ipeirotis, Perona, and Belongie]{7298658}
Grant Van~Horn, Steve Branson, Ryan Farrell, Scott Haber, Jessie Barry, Panos Ipeirotis, Pietro Perona, and Serge Belongie.
\newblock Building a bird recognition app and large scale dataset with citizen scientists: The fine print in fine-grained dataset collection.
\newblock In \emph{2015 IEEE Conference on Computer Vision and Pattern Recognition (CVPR)}, pages 595--604, 2015.

\bibitem[Van~Horn et~al.(2021)Van~Horn, Cole, Beery, Wilber, Belongie, and Mac~Aodha]{van2021benchmarking}
Grant Van~Horn, Elijah Cole, Sara Beery, Kimberly Wilber, Serge Belongie, and Oisin Mac~Aodha.
\newblock Benchmarking representation learning for natural world image collections.
\newblock In \emph{Proceedings of the IEEE/CVF conference on computer vision and pattern recognition}, pages 12884--12893, 2021.

\bibitem[Varghese and Sambath(2024)]{varghese2024yolov8}
Rejin Varghese and M Sambath.
\newblock Yolov8: A novel object detection algorithm with enhanced performance and robustness.
\newblock In \emph{2024 International Conference on Advances in Data Engineering and Intelligent Computing Systems (ADICS)}, pages 1--6. IEEE, 2024.

\bibitem[Wah et~al.(2011)Wah, Branson, Welinder, Perona, and Belongie]{wah2011caltech}
Catherine Wah, Steve Branson, Peter Welinder, Pietro Perona, and Serge Belongie.
\newblock The caltech-ucsd birds-200-2011 dataset.
\newblock 2011.

\bibitem[Wang et~al.(2022)Wang, Xie, Li, Fan, Song, Liang, Lu, Luo, and Shao]{wang2022pvt}
Wenhai Wang, Enze Xie, Xiang Li, Deng-Ping Fan, Kaitao Song, Ding Liang, Tong Lu, Ping Luo, and Ling Shao.
\newblock Pvt v2: Improved baselines with pyramid vision transformer.
\newblock \emph{Computational Visual Media}, 8\penalty0 (3):\penalty0 415--424, 2022.

\bibitem[Wu et~al.(2021)Wu, Xiao, Codella, Liu, Dai, Yuan, and Zhang]{cvt}
Haiping Wu, Bin Xiao, Noel Codella, Mengchen Liu, Xiyang Dai, Lu Yuan, and Lei Zhang.
\newblock Cvt: Introducing convolutions to vision transformers.
\newblock In \emph{Proceedings of the IEEE/CVF international conference on computer vision}, pages 22--31, 2021.

\bibitem[Xie et~al.(2017)Xie, Girshick, Doll{\'a}r, Tu, and He]{xie2017aggregated}
Saining Xie, Ross Girshick, Piotr Doll{\'a}r, Zhuowen Tu, and Kaiming He.
\newblock Aggregated residual transformations for deep neural networks.
\newblock In \emph{Proceedings of the IEEE conference on computer vision and pattern recognition}, pages 1492--1500, 2017.

\bibitem[Zhang et~al.(2022)Zhang, Chen, and Zang]{zhang2022fine}
Yong Zhang, Weiwen Chen, and Ying Zang.
\newblock Fine-grained vision categorization with vision transformer: A survey.
\newblock In \emph{2022 IEEE 8th International Conference on Computer and Communications (ICCC)}, pages 1910--1915. IEEE, 2022.

\bibitem[Zhao et~al.(2017)Zhao, Shi, Qi, Wang, and Jia]{zhao2017pspnet}
Hengshuang Zhao, Jianping Shi, Xiaojuan Qi, Xiaogang Wang, and Jiaya Jia.
\newblock Pyramid scene parsing network.
\newblock In \emph{CVPR}, 2017.

\end{thebibliography}
